\documentclass{bmvc2k}

\title{Towards Reliable AI-Based Histological Staining:
A Systematic Study of Scaling and Uncertainty in Unpaired Generative Models}

\addauthor{Qasim Siddiqui}{qasim.siddiqui@uni-leipzig.de}{1,2}
\addauthor{Adrian Friebel}{friebel@izbi.uni-leipzig.de}{1,2}
\addauthor{Maiju Myllys}{myllys@ifado.de}{3}
\addauthor{Zaynab Hobloss}{hobloss@ifado.de}{3}
\addauthor{Daniela González}{gonzalez@ifado.de}{3}
\addauthor{Ahmed Ghallab}{ghallab@ifado.de}{3,4}
\addauthor{Stefan Hoehme}{hoehme@uni-leipzig.de}{1,2}

\addinstitution{
  Interdisciplinary Centre of Bioinformatics (IZBI) \\
  Universität Leipzig \\
  Leipzig, Germany
}

\addinstitution{
  Center for Scalable Data Analytics and Artificial Intelligence (ScaDS.AI) \\
  University of Leipzig \\
  Leipzig, Germany
}

\addinstitution{
  Department of Toxicology \\
  Technical University Dortmund \\
  Dortmund, Germany
}

\addinstitution{
  Faculty of Veterinary Medicine \\
  Qena University \\
  Qena, Egypt
}

\runninghead{Siddiqui \etal}{Towards Reliable AI-Based Histological Staining}

\usepackage[section]{placeins}

\usepackage{amssymb}
\usepackage{booktabs}
\usepackage{multirow}
\usepackage{stmaryrd}

\def\eg{\emph{e.g}\bmvaOneDot}

\def\etal{\emph{et al}\bmvaOneDot}

\newcommand{\suppsec}[1]{Sec.~S#1}

\newlength{\arxivgutter}
\begin{document}

\raggedbottom

\maketitle

\begin{abstract}
Liver fibrosis, the principal predictor of long-term outcome in chronic liver disease,
is staged from histological estimates of collagen content. Sirius Red~(SR) provides
the standard quantitative readout (collagen proportionate area, CPA) but is not
acquired at every clinical centre and consumes tissue, time, and reagent cost beyond the
routine Hematoxylin and eosin~(H\&E) stain. AI-based virtual staining can generate
SR directly from H\&E, yet systematic benchmarks of unsupervised models are scarce and
their predictive uncertainty has not been quantified, even though visually plausible outputs may
not faithfully reproduce the underlying tissue structure.
We therefore benchmark six unsupervised image-to-image architectures (GAN-based and
diffusion-based) across 54 scaling configurations on a newly released paired
H\&E to SR mouse liver dataset, the first open resource for this translation task.
Each configuration is evaluated jointly on perceptual, distributional, and
task-specific axes plus a blinded expert reader study; the best per family is then
retrained as a deep ensemble, the first systematic comparison of epistemic uncertainty
across unsupervised stain-to-stain architectures.
Across families, perceptual quality, task-specific error, and ensemble agreement
measure largely independent axes of model fitness: GAN-based methods cluster tightly
on perceptual metrics yet differ substantially on task error and ensemble agreement,
while the diffusion-based method (CycleDiffusion) is qualitatively different on all three. No single
metric captures these differences, so reliable virtual staining requires reporting and
selecting on all three jointly. The dataset, tiling pipeline, models, and evaluation
code are released publicly.
\end{abstract}

\section{Introduction}

Histopathology in most cases relies on a panel of stains.
Hematoxylin and eosin~(H\&E), the most widely available routine stain, reveals overall
tissue architecture, while specialised histochemical stains highlight specific tissue
components such as collagen fibres, and immunohistochemical stains visualise
particular proteins via antibody binding to infer cell types or functional states such
as proliferation or necrosis~\cite{lefkowitch2006special, krishna2013role}.
In clinical and preclinical practice, however, each additional stain requires a separate
tissue section. This consumes limited biopsy material~\cite{rivenson2020emerging, dehaan2021deep},
extends laboratory processing time~\cite{dehaan2021deep}, raises per-case reagent and
labour cost (substantially so for antibody-based panels), and introduces spatial
misalignment between stains because tissue deforms during sectioning and
mounting~\cite{gatenbee2023virtual, chen2022mvfstain}.
These costs grow with each additional stain: a full tissue characterisation panel may
require four or more serial sections from a single
biopsy~\cite{clark2017immunohistochemistry}, and because a biopsy yields only a
limited number of usable sections, every extra stain trades against either reserve
material for downstream molecular assays or the breadth of complementary stains that
can be examined on the same specimen.

Liver pathology is a particularly acute case.
Chronic liver disease is staged primarily by the degree of fibrosis, the progressive
deposition of collagen in the extracellular matrix, which is the principal predictor of
long-term outcome.
While fibrosis can be qualitatively appreciated on H\&E, Sirius Red~(SR) provides the
standard \emph{quantitative} readout: SR binds collagen with high specificity and is the
histochemical stain of reference for measuring fibrosis as the collagen proportionate
area~(CPA), correlating more strongly with serum fibrosis biomarkers than the widely
used Masson's trichrome stain~\cite{huang2013sirius}.
Inferring SR directly from a routinely acquired H\&E section would therefore make a
clinically meaningful fibrosis readout available wherever H\&E is, without consuming
additional tissue or laboratory resources.

\begin{figure*}[t!]
  \centering
  \includegraphics[width=0.90\linewidth]{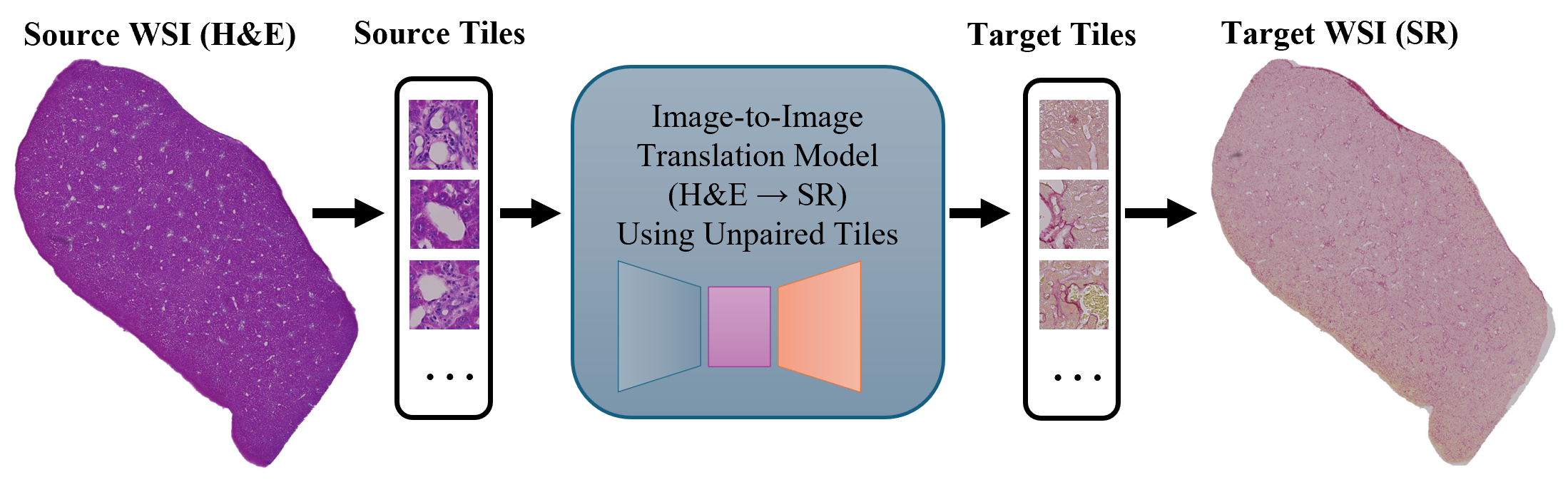}
  \caption{%
    Overview of the virtual H\&E~$\to$~SR staining pipeline studied in this work.
    A whole-slide image~(WSI) acquired with H\&E staining is decomposed into
    $256{\times}256$\,pixel tiles, which are passed through an unpaired
    image-to-image~(I2I) translation model to produce virtual Sirius Red tiles.
    The translated tiles are reassembled into a virtual SR WSI, from which
    collagen proportionate area~(CPA) can be quantified without requiring a
    second, physical SR-stained section from the same tissue block.
  }
  \label{fig:pipeline_overview}
\end{figure*}

Virtual staining offers an attractive solution: it uses deep learning to predict a
target stain directly from an existing H\&E image
(Figure~\ref{fig:pipeline_overview}), and additionally enables retrospective analysis of
existing H\&E archives without acquiring new
sections~\cite{rivenson2020emerging, latonen2024virtual}.
Supervised approaches, however, require pixel-aligned image pairs from adjacent tissue sections,
which is impractical at scale and introduces bias from tissue
deformation~\cite{aatresh2024virtual}.
\textit{Unsupervised} methods, which learn from unpaired image collections, are better
suited to this setting. However, no prior work has systematically compared them for
H\&E~$\to$~SR translation in liver tissue; existing comparisons focus almost entirely on
H\&E~$\to$~IHC or on non-liver
tissue~\cite{visgab2025, adib2024dual, klockner2025he}.
The existing benchmarks fix a single model scale and data budget~\cite{visgab2025, klockner2025he},
leaving open how model capacity and training data jointly affect translation quality, an
important question because liver biopsy cohorts vary widely in size across clinical centres.

Beyond translation quality, we also need a signal of when a virtual SR output can be
trusted. Task-specific evaluation can verify aggregate translation quality at the WSI
level (Section~\ref{sec:metrics}), but in unpaired settings no ground-truth SR is available for any given
H\&E tile, so individual tile outputs cannot be checked against truth even though
clinical decisions may hinge on local features within a single tile.
Training ensembles of independently seeded models and comparing their outputs reveals where
they converge on the same answer and where they do not on tile level: disagreement flags inputs for
which the unsupervised objective does not pin down a unique translation, providing an
indicator of epistemic uncertainty that complements task-specific evaluation.

Consequently, we present three main results that advance unsupervised stain
translation toward more reliable and uncertainty-aware evaluation.

First, to map where translation quality actually lies in the design space of model
capacity and training-data budget, we conduct a full-factorial scaling study of six
unsupervised I2I architectures spanning GAN-based and diffusion-based families
(CycleGAN~\cite{zhu2017unpaired}, UNIT~\cite{liu2017unsupervised},
MUNIT~\cite{huang2018multimodal}, DCLGAN~\cite{han2021dual},
UVCGAN~\cite{torbunov2022uvcgan}, and CycleDiffusion~\cite{wu2022cyclediffusion}) at
three generator capacities (${\sim}10$, ${\sim}50$, ${\sim}100$~million parameters) and
three data fractions (25\%, 50\%, 100\%). Each of the 54 configurations is evaluated
using a task-specific CPA mean absolute error~(CPA~MAE), computed by applying a frozen
collagen segmentation model to real and virtual SR images, alongside three perceptual metrics: FID (distributional
realism), LPIPS (perceptual similarity), and patch-SSIM (local structural similarity).
Reporting all four metrics jointly exposes where perceptual rankings agree with
task-level accuracy and where they diverge.

Building on this study, to quantify where these models are committed to a single
answer and where they are not, we train deep ensembles of independently seeded
models for the best configuration per family, producing spatially resolved uncertainty
maps. We provide the first systematic comparison of epistemic uncertainty across unsupervised
stain-to-stain architectures.

\section{Related Work}
\label{sec:related_work}

\noindent\textbf{AI-based Histological Staining.}
In liver histopathology specifically, an unsupervised GAN validated on $n{=}574$ NASH
WSIs~\cite{kieffer2020largescale} produced virtual H\&E~$\to$~trichrome translations
that pathologists could not reliably distinguish from real images, and a prior-guided
GAN~\cite{li2023unpaired} reached a Spearman correlation of $r_s{=}0.82$ between virtual
and real fibrosis scores on the same task.
Recent benchmarks covering a broader set of models and stains~\cite{visgab2025,
staindiffuser2024, klockner2025he} consistently find that quantitative rankings disagree
with pathologists' visual assessments.

\noindent\textbf{Evaluation Frameworks for Virtual Staining.}
Khan~\etal~\cite{khan2023effect} showed directly that standard image quality metrics do
not reliably reflect whether virtually stained H\&E images retain the histological structures
and cellular morphology necessary for pathological assessment, a finding that
motivates the evaluation design adopted here and the broader argument for
pathology-aware evaluation.
A growing body of work therefore argues for \textit{task-specific evaluation}: applying
downstream clinical tasks to virtual images to verify that diagnostically relevant
content is preserved~\cite{renal_seg2025, her2task2024}.
Task-specific evaluation, however, only verifies aggregate WSI-level translation quality and
cannot flag where individual tile outputs fail; pairing it with model-level uncertainty
estimation would complete the picture, yet no comparable uncertainty framework exists
for unpaired stain translation.

\noindent\textbf{Epistemic Uncertainty in Virtual Staining and I2I Translation.}
Epistemic uncertainty captures the component of predictive uncertainty attributable to
limited training data or model capacity. It is most commonly quantified via Monte
Carlo~(MC) dropout~\cite{gal2016dropout}, which approximates Bayesian posterior sampling
through repeated stochastic forward passes, or via deep
ensembles~\cite{lakshminarayanan2017simple}, which measure disagreement across
independently trained models.
In digital histopathology, both approaches have been applied to H\&E classification and
segmentation, enabling high-confidence predictions and automatic rejection of ambiguous
cases~\cite{uncertainty_informed_hist2022}.
Yet within the I2I translation literature, uncertainty-awareness in the \textit{unpaired}
setting remains underexplored.
The closest prior work is UGAC~\cite{ugac2021}, which extended CycleGAN with a
generalised Gaussian cycle-consistency loss to produce per-pixel epistemic uncertainty
estimates for MRI modality translation; UGAC reported that the resulting per-pixel
variance correlates with translation error and stabilises training under noisy
supervision.
That setting differs from ours in two ways: the target domain is MRI rather than
histology, and uncertainty is encoded into the loss as a Gaussian likelihood rather than
measured ex post via ensemble disagreement.
In supervised virtual cell staining, ensemble-based translation produced uncertainty
estimates with Pearson $\rho{\approx}0.83$ against per-tile translation
error~\cite{insilicolabeling2023}.
No prior study has applied MC~dropout or deep ensembles to quantify epistemic
uncertainty in unpaired GAN- or diffusion-based \textit{histological stain} translation,
leaving open which image regions are most uncertain and how informative the resulting
ensemble disagreement is in practice.
\section{Dataset}
\label{sec:dataset}

The dataset comprises 70 whole-slide images of mouse liver tissue
(35 H\&E and 35 SR sections) obtained in a bile-duct ligation experiment of cholestatic
liver disease~\cite{ghallab2025asbt}, covering four post-surgical disease stages
(days~3, 21, 42, and 63, plus sham controls) from minimal to advanced hepatic fibrosis.
For each of 35 C57BL/6N mice, serial sections from the same tissue block were stained
with H\&E and SR. Because the two sections are non-adjacent cuts, the dataset is
inherently \emph{unpaired}.
WSIs were acquired at 20-fold magnification and tiled into non-overlapping
$512{\times}512$\,pixel patches. The tiles were downsampled to $256{\times}256$\,pixels
to match the input size of all six evaluated baselines and to lower per-batch GPU
memory at training time, yielding an effective resolution of $0.442\,\mu$m/pixel.
30 animals (60 WSIs) contribute 201{,}257 H\&E and 196{,}415 SR training tiles, with
the remaining 5 animals (10 WSIs, 16{,}290 tiles per domain) reserved for testing;
the split is partitioned at the animal level to prevent data leakage.
The held-out animals cover sham controls and days~3, 21, and 42; the one day-63
specimen could not be co-registered across its two stains and was excluded, so the
most advanced fibrosis stage is absent from the test set.
This constitutes the first open resource for unsupervised H\&E~$\to$~SR translation in
mouse liver tissue; we release the 70 WSIs together with the tiling
and preprocessing pipeline used to generate the $256{\times}256$ tiles and the full
evaluation code (CPA~MAE, FID, LPIPS, patch-SSIM, and ensemble variance).
Animal experiments were approved by LANUV, North Rhine-Westphalia, Germany
(application 81-02.04.2022.A286)~\cite{ghallab2025asbt}. Full
acquisition, staining, and preprocessing details are in \suppsec{1}.
%
%
%

\section{Models}
\label{sec:models}

We evaluate six unsupervised I2I architectures (schematic in \suppsec{2}):
CycleGAN~\cite{zhu2017unpaired}, the foundational unpaired translation framework,
together with five widely cited methods that each extend it along a distinct
methodological direction.
DCLGAN~\cite{han2021dual} adds a dual patch-level contrastive loss between domains for
finer-grained content alignment.
UNIT~\cite{liu2017unsupervised} and MUNIT~\cite{huang2018multimodal} both share a latent
space across domains but realise it differently: UNIT through a shared VAE bottleneck between
the domain generators, while MUNIT injects per-domain content and style codes (style via
Adaptive Instance Normalisation) into the same latent space.
UVCGAN~\cite{torbunov2022uvcgan} retains the cycle objective but replaces the ResNet
bottleneck with a Vision Transformer block preceded by masked-image pretraining, aiming
for improved stability at higher capacity.
CycleDiffusion~\cite{wu2022cyclediffusion} departs from the GAN paradigm entirely by
coupling the domains in a shared diffusion noise space accessed via DDIM~\cite{song2021ddim}
inversion and sampling, replacing adversarial training with a deterministic generative
process.
Each model is trained at three generator capacities (${\sim}10$, ${\sim}50$, and
${\sim}100$~million parameters) to measure how translation quality responds to added
capacity within each architecture family.
Although pre-trained priors and noise-inversion methods can improve fidelity, we train
all models from scratch under a fixed data and compute budget so that capacity and data
effects are compared fairly across families.
Full architectural details and loss formulations are in \suppsec{2}.

\section{Scaling Study: Experimental Design and Results}
\label{sec:experiments}



\textbf{Training Protocol.}
All models are trained with the Adam optimiser~\cite{kingma2015adam} (full hyperparameters
in \suppsec{3}). We count training in optimiser steps rather than epochs,
ensuring fair comparison across data-fraction configurations with different tile counts.
CycleGAN, UNIT, MUNIT, DCLGAN, and CycleDiffusion are each trained for $750{,}000$~steps.
UVCGAN follows a two-stage schedule: $250{,}000$~steps of masked-image pretraining
followed by $500{,}000$~steps of cycle-consistent finetuning initialised from the
Stage~1 checkpoint.

\noindent\textbf{Scaling Study Design.}
We conduct a full factorial experiment over the six architectures of
Section~\ref{sec:models}, three generator capacity scales, and three training data
fractions.

\noindent\textbf{Generator capacity.}
Each model is trained at three generator sizes: Small~(S, ${\sim}10$~million parameters),
Medium~(M, ${\sim}50$~million), and Large~(L, ${\sim}100$~million), by adjusting the
base channel width and depth (or equivalent per model family; see \suppsec{2}).
All size variants within a family share the same training protocol, data, and random
seed, so any difference reflects capacity alone.

\noindent\textbf{Training data fraction.}
The training set is partitioned into three nested subsets comprising 25\%, 50\%, and
100\% of the 30 training paired specimens (60 WSIs), constructed via contiguous WSI ranges
to avoid data leakage.
Smaller subsets are strict subsets of larger ones, so differences across data fractions
are not confounded by variation in which slides are included.
The same fixed test set of $N{=}5$ held-out paired specimens (10 WSIs) is used for
evaluation across all configurations.
Together, these axes yield $6\text{ models} \times 3\text{ sizes} \times 3\text{ data
fractions} = \mathbf{54}$ experimental configurations, each trained and evaluated
independently.

\noindent\textbf{Inference.}
At test time, all GAN-based models perform a single forward pass through the trained
$G_{A \to B}$ generator per tile.
CycleDiffusion requires two sequential DDIM passes per tile~\cite{song2021ddim}: the
source-domain DDPM encodes the source tile into the shared noise latent via DDIM
inversion over $T_{\text{encode}}{=}$200~steps, and the target-domain DDPM then decodes
it into the target domain via DDIM sampling over $T_{\text{sample}}{=}$200~steps.

%
%


\subsection{Evaluation Metrics}
\label{sec:metrics}

Since the ultimate objective of virtual staining is to support downstream biological
analysis, task-specific evaluation provides the most direct measure of model performance.
A translation that faithfully preserves the relevant tissue signal will produce
biologically consistent quantitative readouts when the same analysis pipeline is applied
to both real and virtual images.
We supplement this with structural, perceptual, and distributional measures that
characterise the visual quality of the generated stains.

Because H\&E and SR sections are non-adjacent cuts (Sec.~\ref{sec:dataset}), direct
pixel-level comparison cannot serve as ground truth.
We therefore co-register the test-set H\&E and SR WSIs with the VALIS feature-based
framework~\cite{gatenbee2023virtual}, which reaches tile-scale but not pixel-exact
correspondence, so residual misalignment inflates apparent error for any pixel-level
comparison metric.

\noindent\textbf{Perceptual and Distributional Metrics:} Given these constraints, we report patch-based SSIM~\cite{wang2004ssim},
LPIPS~\cite{zhang2018lpips}, and FID~\cite{heusel2017fid}; each captures a different
aspect of perceptual or distributional quality and each has known limitations for
virtual staining.
Patch-based SSIM computes structural similarity over local patches to reduce
sensitivity to residual misregistration, and LPIPS measures VGG-based deep-feature
distance over the co-registered tile pairs without strict pixel pairing.
FID compares real and generated SR distributions using InceptionV3 features, capturing
population-level realism but unable to detect localised semantic failure on individual
tiles.
Because these InceptionV3 features are trained on natural images rather than histology,
FID can reward distributional realism while missing collagen-specific structure; we
therefore treat CPA~MAE, not FID, as the primary metric.

\noindent\textbf{Task-Specific Evaluation: Collagen Proportionate Area (CPA) Agreement.}
Although the two sections cannot be compared pixel-by-pixel, they share the same
underlying collagen content and therefore allow a paired comparison at the level of
biological quantity. To realise this, a frozen nnU-Net~v2~\cite{isensee2021nnunet}
collagen segmentation model~\cite{deng2026} trained on real SR WSIs is applied
identically to (i)~the real SR test WSIs and (ii)~the virtual SR WSIs produced by each
model configuration.
Deep learning segmentation models of this kind reach Dice scores of $0.83$--$0.94$
on Sirius Red~\cite{liverfibrosis2025, deng2026}, demonstrating that automated CPA
measurement is clinically viable.
For each WSI, CPA is computed as the fraction of tissue pixels labelled as SR-positive
collagen (background excluded).
CPA is the standard quantitative readout for hepatic fibrosis assessment and is
directly comparable across slides~\cite{huang2013sirius}.

The primary task-specific metric is the \textit{mean absolute error}~(MAE) between
paired real-SR and virtual-SR CPA values:
\begin{equation}
  \mathrm{MAE}_{\mathrm{CPA}} = \frac{1}{N}\sum_{i=1}^{N}
  \bigl|\mathrm{CPA}^{\mathrm{virtual}}_{i} - \mathrm{CPA}^{\mathrm{real}}_{i}\bigr|,
  \label{eq:cpa_mae}
\end{equation}
where $N$ is the number of held-out paired specimens.
A lower CPA~MAE indicates that the virtual stain supports collagen quantification
consistent with the real stain, serving as the primary biologically grounded benchmark
for model evaluation in this study.

\noindent\textbf{Expert Reader Study.}
Automated metrics are complemented by a blinded reader study with three expert readers,
two clinical liver experts and one reader experienced with histological images, conducted on the
held-out set for the study-best and study-worst configurations.
Readers first decided whether each SR tile was real or generated, then rated whether the
collagen in a generated tile is biologically plausible given its matched H\&E input on a
forced-choice four-point scale.
The full protocol is given in \suppsec{6}.

\subsection{Results: Scaling Study}
\label{sec:results_scaling}

\begin{figure*}[t!]
  \centering
  \includegraphics[width=\linewidth]{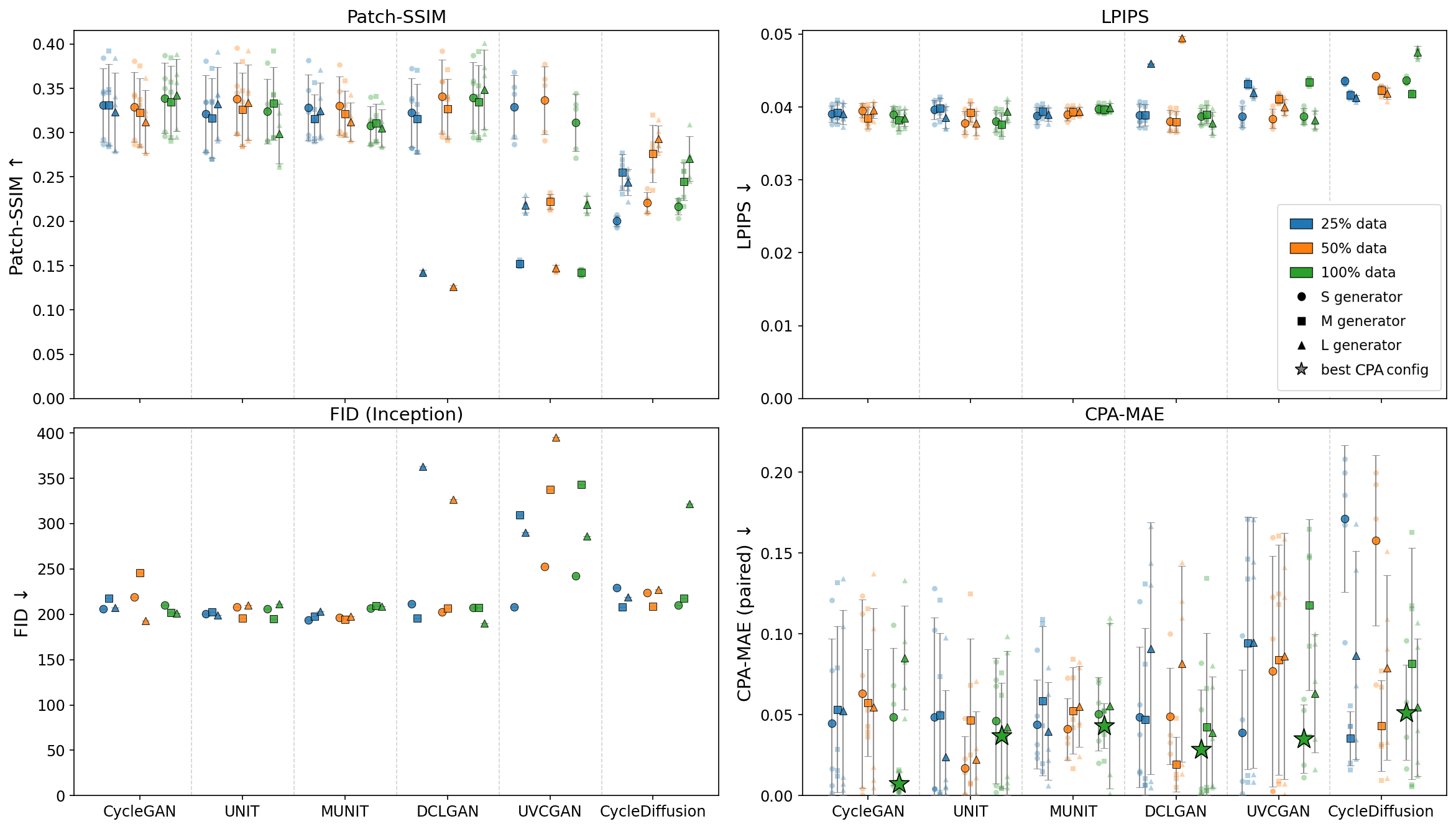}
  \caption{%
    All four evaluation metrics across the 54 scaling configurations
    ($6~\text{models} \times 3~\text{sizes} \times 3~\text{data fractions}$).
    Colour encodes training data fraction (blue: 25\%, orange: 50\%, green: 100\%);
    marker shape encodes generator size (circle: small~S, square: medium~M, triangle: large~L).
    Large star markers indicate the best-performing model size at the 100\% data fraction
    per model family, selected for ensemble uncertainty estimation in
    Section~\ref{sec:uncertainty}.
    Error bars on Patch-SSIM, LPIPS, and CPA~MAE denote $\pm$\,std over $N{=}5$ held-out
    paired specimens (10 WSIs); FID is a set-level statistic with no per-sample variance.
    Vertical axis directions: Patch-SSIM higher is better; LPIPS, FID, CPA~MAE lower
    is better.%
  }
  \label{fig:results_overview}
\end{figure*}

Quantitative results for all 54 configurations are shown in
Figure~\ref{fig:results_overview}, which plots all four metrics jointly across the
full experiment.
Each point corresponds to one configuration; colour encodes data fraction and marker
shape encodes generator size.
Four high-level patterns emerge: CPA~MAE spans over an order of magnitude
($0.008$--$0.171$) while the perceptual metrics saturate within a narrow band for
well-behaved GAN configurations; CycleDiffusion attains systematically lower Patch-SSIM
irrespective of scale; two model families exhibit training instability under specific
scaling conditions; and perceptual metric rankings correlate only moderately with the
CPA~MAE ranking.

\begin{table}[t]
\centering
\caption{Best configuration per model family at 100\% training data, selected on
lowest CPA~MAE (the rule also used for ensemble uncertainty estimation in
Section~\ref{sec:uncertainty}). \textbf{Bold} marks the best value per metric
across families. Patch-SSIM higher is better; LPIPS, FID, and CPA~MAE lower is
better. Patch-SSIM, LPIPS, and CPA~MAE are mean~$\pm$~std over the $N{=}5$
held-out paired specimens; FID is a set-level statistic.}
\label{tab:best_per_family}
\small
\begin{tabular}{llcccc}
\toprule
Family & Size & Patch-SSIM~$\uparrow$ & LPIPS~$\downarrow$ & FID~$\downarrow$ & CPA~MAE~$\downarrow$ \\
\midrule
CycleGAN       & M & $0.334 \pm 0.041$         & $\mathbf{0.038} \pm 0.001$ & $201.8$           & $\mathbf{0.008} \pm 0.007$ \\
UNIT           & M & $0.333 \pm 0.041$         & $\mathbf{0.038} \pm 0.002$ & $\mathbf{195.2}$  & $0.037 \pm 0.033$ \\
MUNIT          & M & $0.310 \pm 0.022$         & $0.040 \pm 0.001$         & $209.5$           & $0.043 \pm 0.014$ \\
DCLGAN         & S & $\mathbf{0.339} \pm 0.040$ & $0.039 \pm 0.001$         & $207.4$           & $0.028 \pm 0.037$ \\
UVCGAN         & S & $0.311 \pm 0.032$         & $0.039 \pm 0.001$         & $242.5$           & $0.035 \pm 0.021$ \\
CycleDiffusion & S & $0.217 \pm 0.009$         & $0.044 \pm 0.000$         & $210.4$           & $0.051 \pm 0.029$ \\
\bottomrule
\end{tabular}
\end{table}

Table~\ref{tab:best_per_family} reports the best configuration per family at 100\%
training data; the per-family results across the full size~$\times$~data-fraction
grid are summarised below, and complete metric tables for all 54 configurations are
in \suppsec{4}.

\paragraph{CycleGAN.}
Patch-SSIM ($0.312$--$0.342$) and LPIPS ($0.038$--$0.040$) are near-invariant across
all nine configurations; FID ($193$--$245$) shows no monotonic size trend.
CPA~MAE is the most discriminative axis: at 100\% data, small/medium/large yield
$0.049$/$0.008$/$0.085$, with the medium generator achieving the lowest CPA~MAE in the
entire study ($0.008$).
Across data fractions, perceptual metrics are insensitive to data volume while CPA~MAE
shows no consistent trend for any generator size.

\noindent\textbf{UNIT.}
Perceptual metrics are stable across the full grid.
CPA~MAE at 100\% data is $0.046$/$0.037$/$0.043$ (S/M/L), with no consistent size
advantage; the small generator achieves its best CPA~MAE at 50\% data ($0.017$) rather
than 100\%.

\noindent\textbf{MUNIT.}
The most stable family: all four metrics span narrow ranges across all nine cells, with
CPA~MAE confined to $0.040$--$0.059$ and no generator size or data fraction yielding a
systematic advantage.

\noindent\textbf{DCLGAN.}
At 100\% data all three sizes are stable (CPA~MAE $0.028$--$0.042$, FID $190$--$208$);
the large generator achieves the study-best FID ($190.0$).
At 50\% and 25\% data, the large generator collapses~(Patch-SSIM $0.126$--$0.143$, FID
$327$--$363$, CPA~MAE $0.082$--$0.091$), while small and medium sizes remain stable
(CPA~MAE $0.019$--$0.049$) at all data fractions.

\noindent\textbf{UVCGAN.}
The small generator is stable across all data fractions (Patch-SSIM $0.311$--$0.337$,
CPA~MAE $0.035$--$0.077$).
Medium and large variants are degraded across all data fractions (Patch-SSIM
$0.142$--$0.222$, FID $287$--$395$, CPA~MAE $0.063$--$0.118$), with the medium
generator reaching its highest CPA~MAE at 100\% data ($0.118 \pm 0.053$).

\noindent\textbf{CycleDiffusion.}
Consistently lower Patch-SSIM ($0.200$--$0.293$) and higher LPIPS ($0.041$--$0.048$)
than all GAN-based families, reflecting the DDIM inversion inductive bias.
At 100\% data, CPA~MAE is $0.051$/$0.082$/$0.055$ (S/M/L); the small generator's
CPA~MAE rises sharply with data reduction (exceeding $0.15$ at both 50\% and 25\% data,
reaching $0.171$ at 25\%), while the medium generator achieves its lowest CPA~MAE at
25\% data ($0.035$) rather than at 100\% ($0.082$).

\begin{figure*}[t!]
  \centering
  \setlength{\tabcolsep}{2pt}
  \renewcommand{\arraystretch}{0.4}
  \small
  \begin{tabular}{lcccc}
     & \textbf{H\&E input} & \textbf{Real SR} & \textbf{CycleGAN-M$\times$100\%} & \textbf{CycleDiff-S$\times$25\%} \\[1pt]
    \textbf{Patch 1} &
    \includegraphics[width=0.20\linewidth]{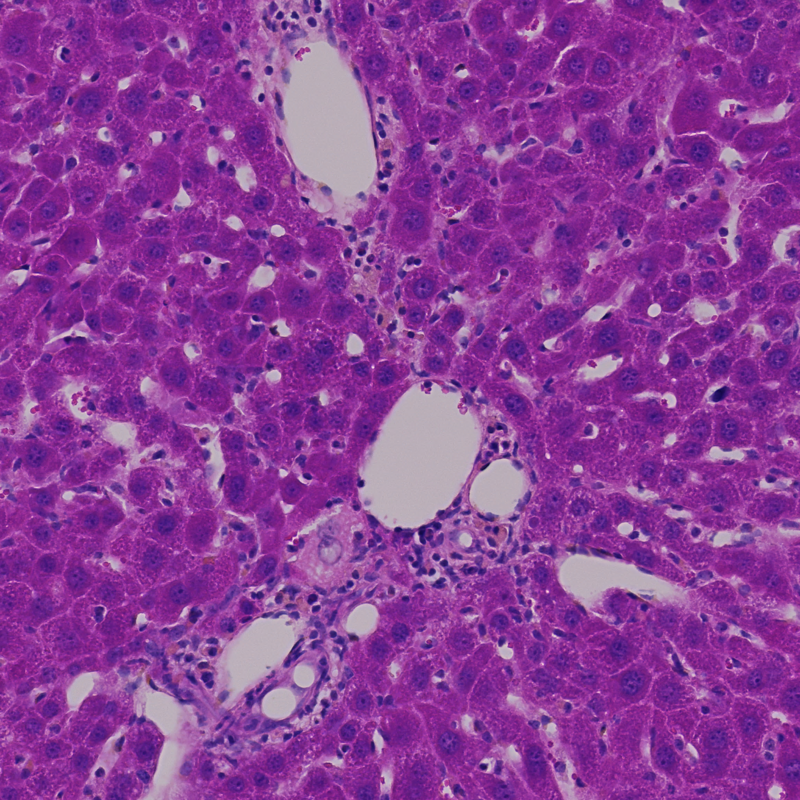} &
    \includegraphics[width=0.20\linewidth]{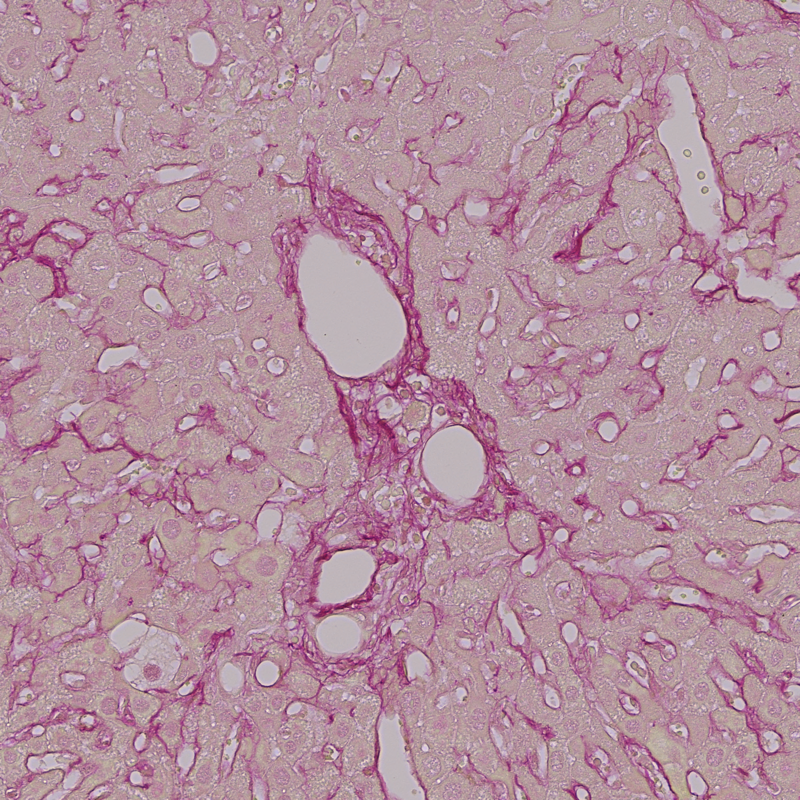} &
    \includegraphics[width=0.20\linewidth]{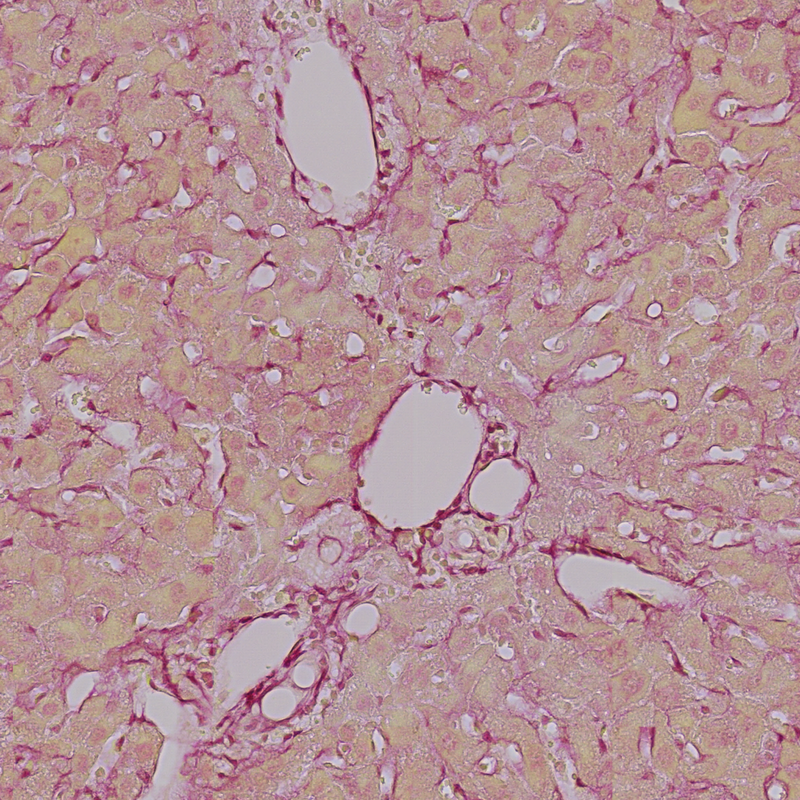} &
    \includegraphics[width=0.20\linewidth]{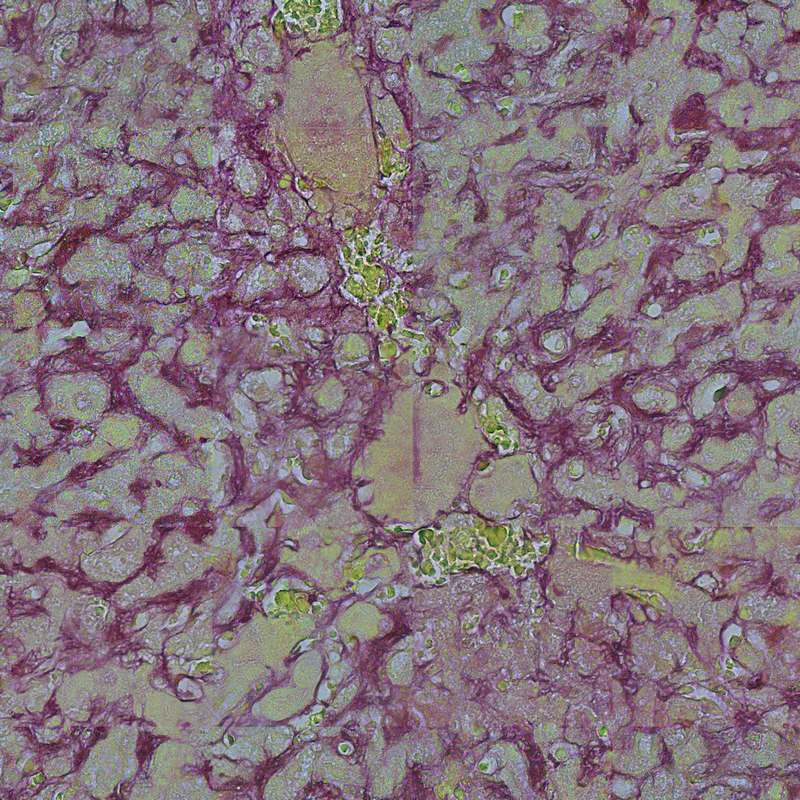} \\
    & &
    \includegraphics[width=0.20\linewidth]{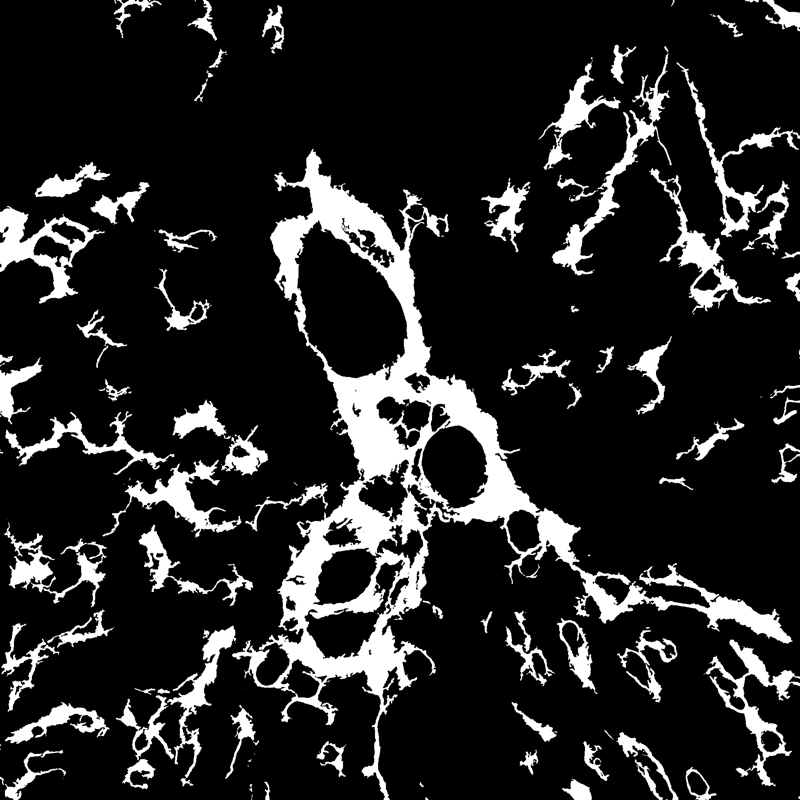} &
    \includegraphics[width=0.20\linewidth]{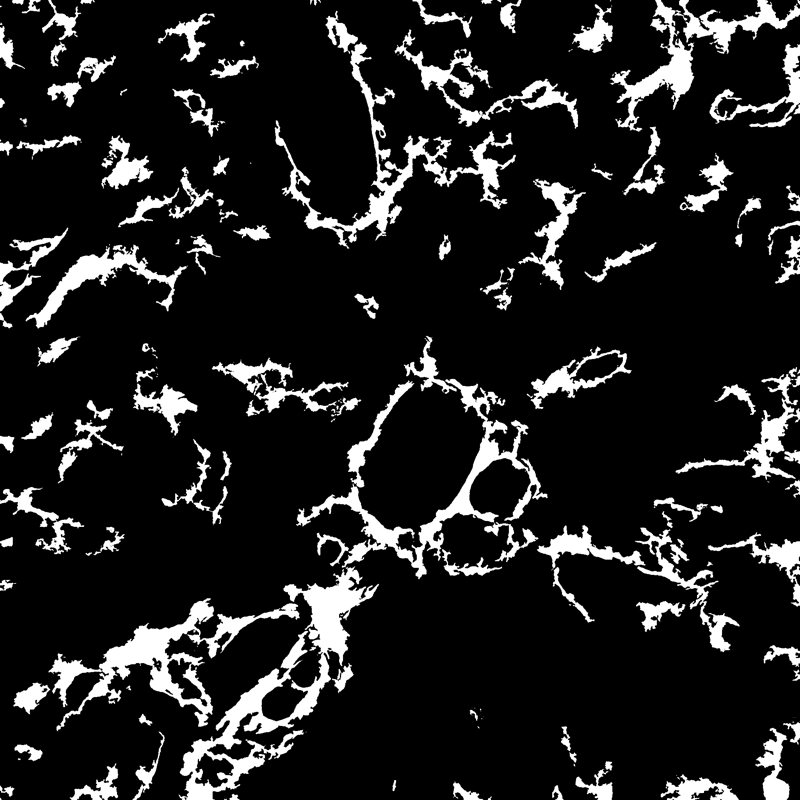} &
    \includegraphics[width=0.20\linewidth]{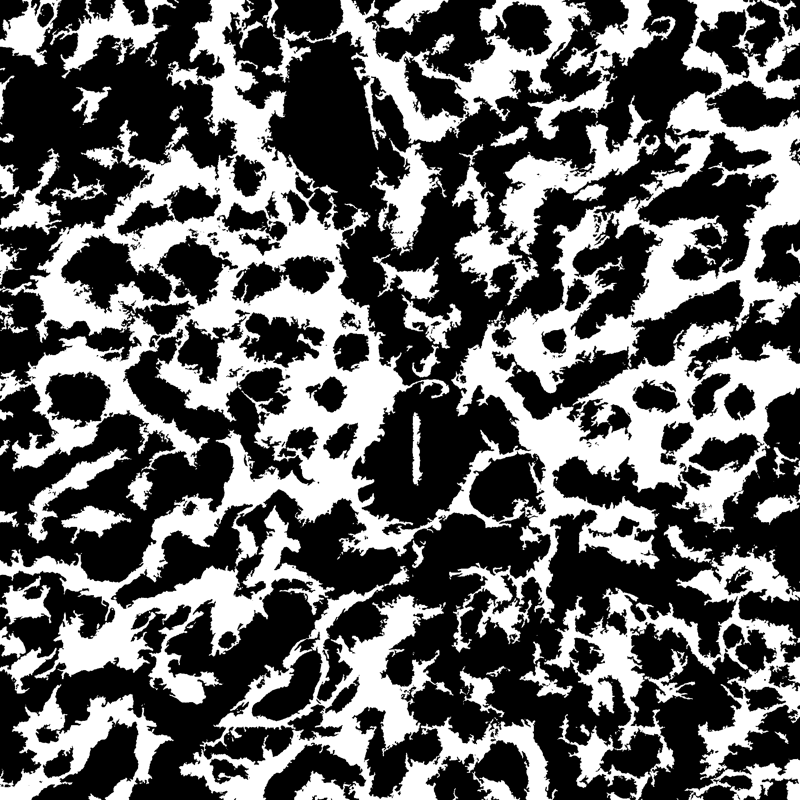} \\[4pt]
    \textbf{Patch 2} &
    \includegraphics[width=0.20\linewidth]{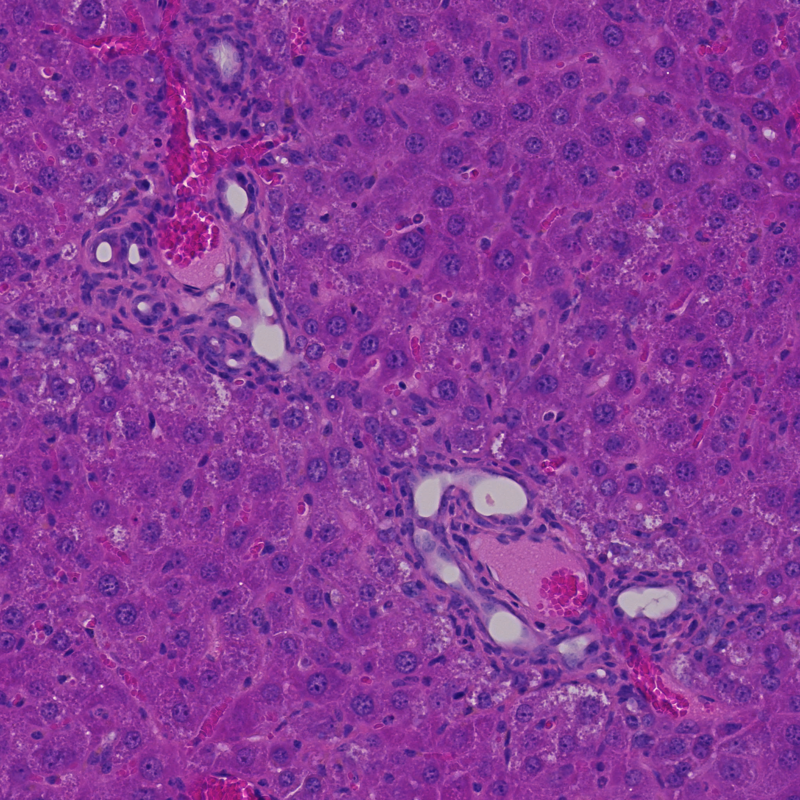} &
    \includegraphics[width=0.20\linewidth]{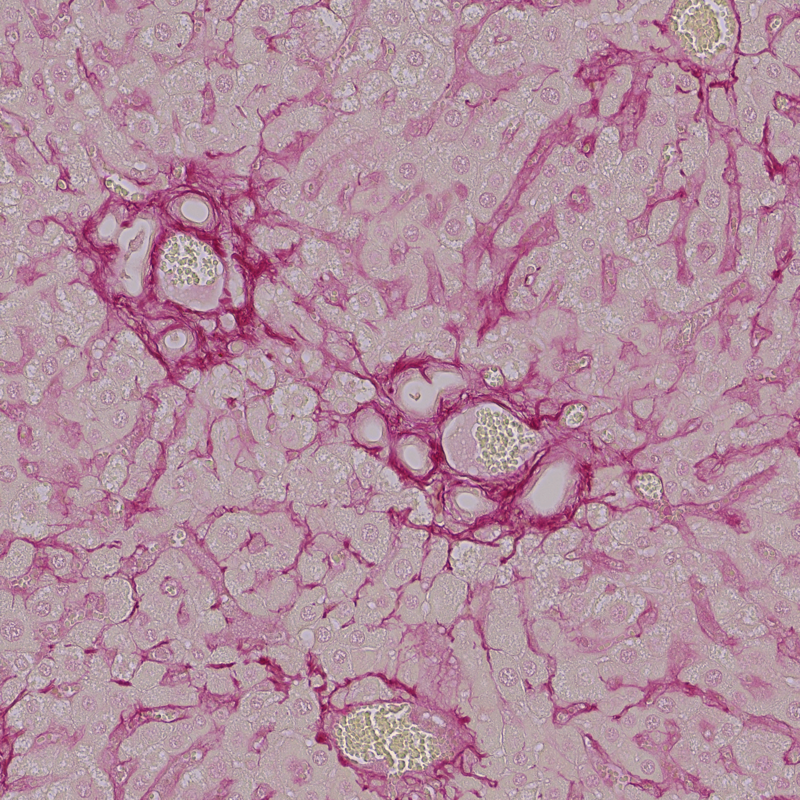} &
    \includegraphics[width=0.20\linewidth]{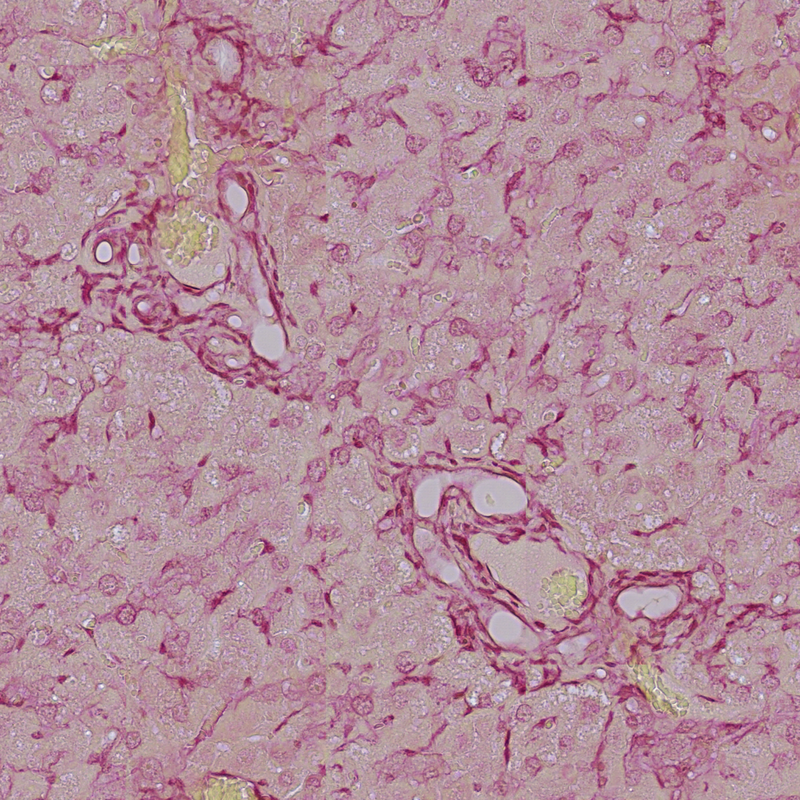} &
    \includegraphics[width=0.20\linewidth]{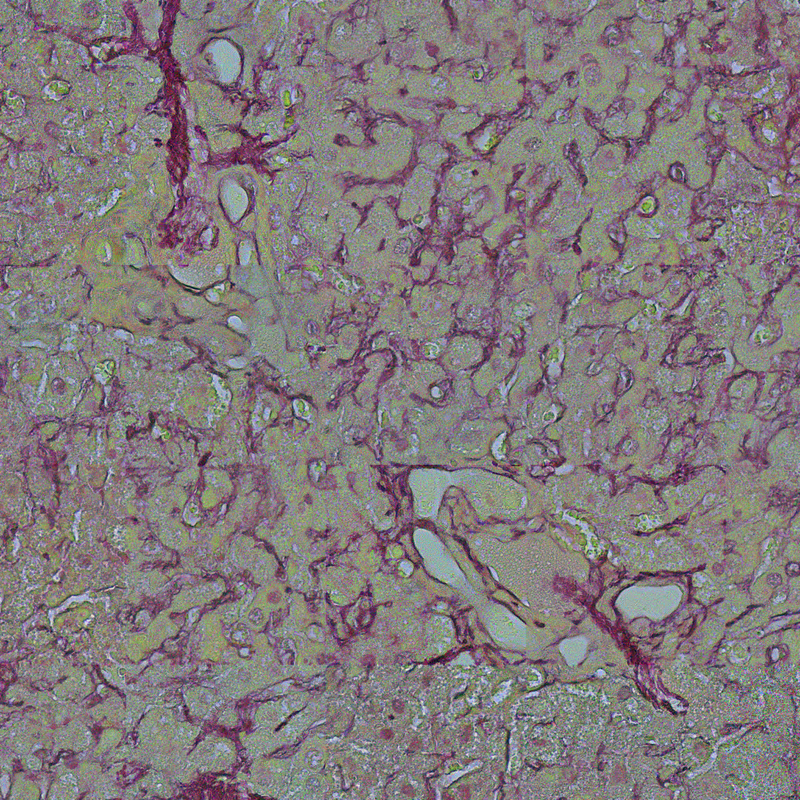} \\
    & &
    \includegraphics[width=0.20\linewidth]{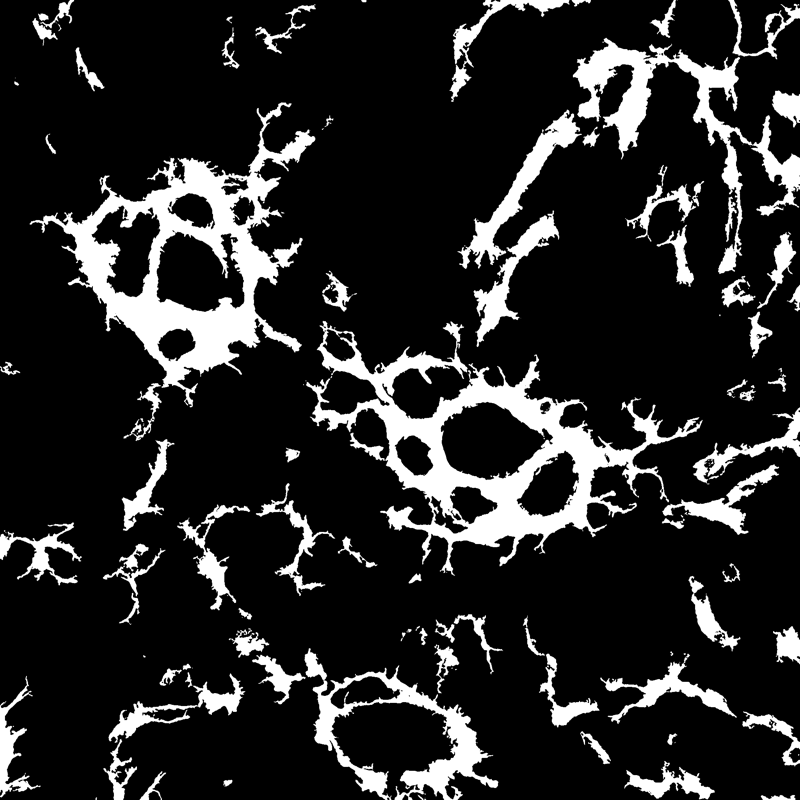} &
    \includegraphics[width=0.20\linewidth]{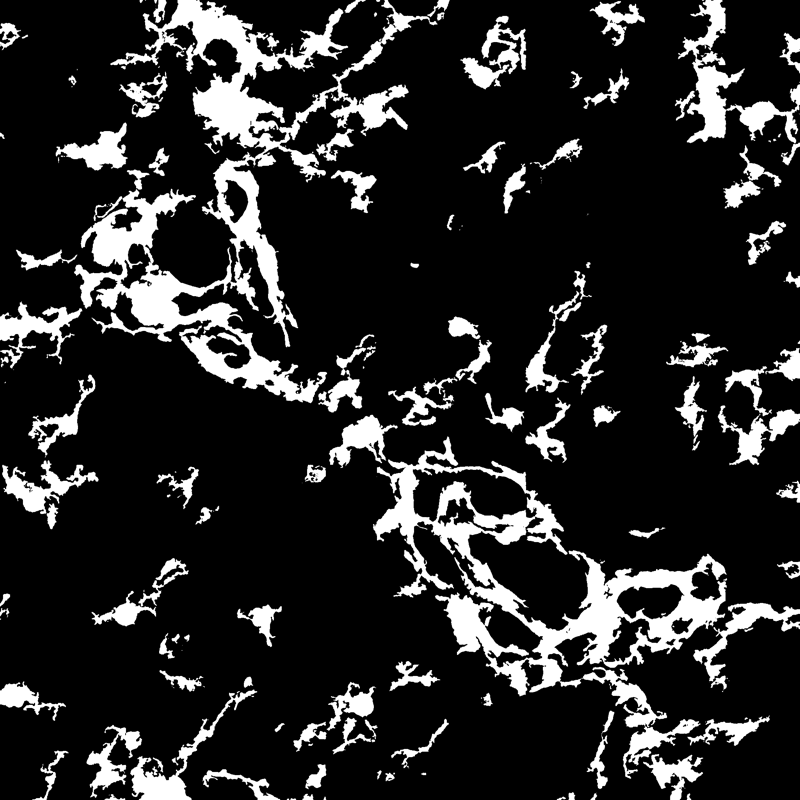} &
    \includegraphics[width=0.20\linewidth]{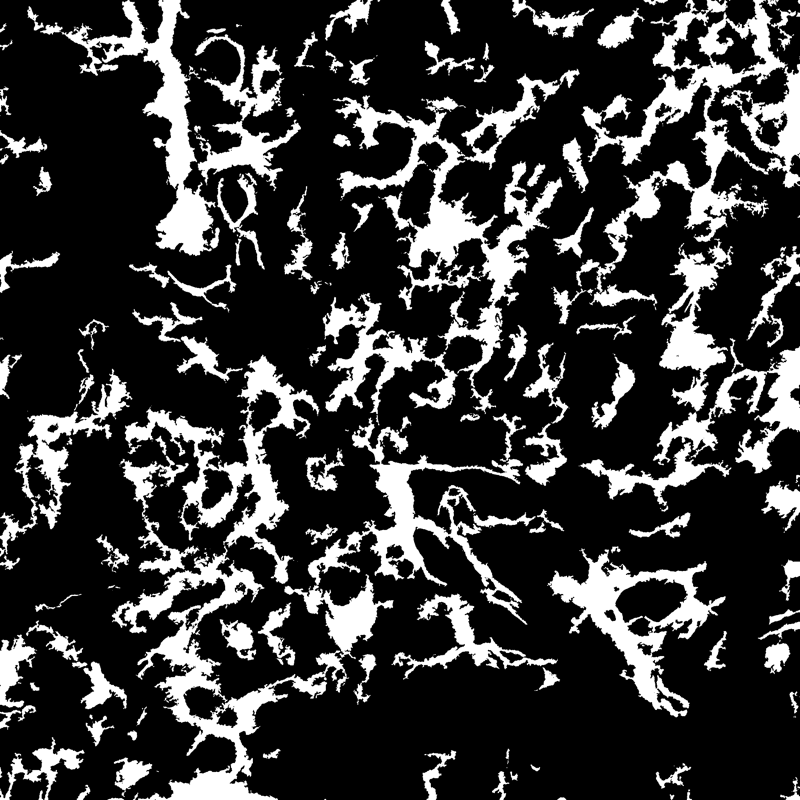} \\
  \end{tabular}
  \caption{%
    Qualitative comparison on two held-out H\&E patches. For each patch, the top row
    shows the H\&E input, the loosely registered real Sirius Red ground truth, and the
    virtual SR produced by CycleGAN's medium generator at 100\% training data and
    CycleDiffusion's small generator at 25\% training data. The bottom row shows the
    nnU-Net~v2 collagen segmentation masks of the corresponding real and virtual SR
    images, from which CPA is computed. Because the H\&E and SR sections are
    non-adjacent cuts, visible differences between the real and virtual SR images
    reflect both true translation error and irreducible inter-section variation.
    These two configurations bracket the study: the CycleGAN-M 100\% configuration
    records the study-best CPA~MAE across all 54 configurations, while the
    CycleDiffusion-S 25\% configuration records the study-worst.%
  }
  \label{fig:qualitative_examples}
\end{figure*}

Figure~\ref{fig:qualitative_examples} shows representative tiles from the study-best
(CycleGAN-M, 100\%~data) and study-worst (CycleDiffusion-S, 25\%~data) configurations
to make these perceptual and task-error contrasts visible.


\begin{table}[h!]
  \centering
  \caption{Spearman rank correlation ($r_s$) between the four metric orderings over
    all 54 configurations (rank~1 = best per metric, so each metric is oriented in its
    own better-is-first direction and no $\uparrow$/$\downarrow$ applies). All
    off-diagonal values significant at $p{<}0.001$.}
  \label{tab:ranking_correlation}
  \begin{tabular}{lcccc}
    \toprule
     & Patch-SSIM & LPIPS & FID & CPA~MAE \\
    \midrule
    Patch-SSIM & 1.00 & & & \\
    LPIPS      & 0.82 & 1.00 & & \\
    FID        & 0.60 & 0.49 & 1.00 & \\
    CPA~MAE    & 0.63 & 0.64 & 0.55 & 1.00 \\
    \bottomrule
  \end{tabular}
\end{table}

\noindent\textbf{Metric agreement.}
Table~\ref{tab:ranking_correlation} shows Spearman rank correlations between the four
metric orderings over all 54 configurations (rank~1 assigned to the best-performing
configuration per metric; all pairs significant at $p{<}0.001$).
Patch-SSIM and LPIPS rankings agree most strongly with each other ($r_s{=}0.82$).
FID shows only moderate agreement with Patch-SSIM ($r_s{=}0.60$) and weak agreement with LPIPS
($r_s{=}0.49$).
CPA~MAE likewise shows only moderate agreement with all three perceptual metrics:
$r_s{=}0.63$ with Patch-SSIM, $r_s{=}0.64$ with LPIPS, and $r_s{=}0.55$ with FID.

\noindent\textbf{Expert reader study.}
The blinded reader study (Section~\ref{sec:metrics}) separates the same two
configurations that bracket the CPA~MAE range.
In the detection task the three readers correctly identified only $38\%$ of
best-configuration tiles as generated, against $62\%$ for the worst configuration, so
the best configuration is not reliably distinguishable from real SR while the worst is.
In the plausibility task all three readers independently rated the best configuration as
plausible ($2.7$, $3.2$, and $3.7$ of $4$) and the worst as implausible ($1.1$, $1.4$,
and $1.7$).
The separation tracks the CPA~MAE ranking ($0.008$ vs.\ $0.171$) rather than the
perceptual scores, which place the two configurations far closer together.
Full protocol and per-reader results are in \suppsec{6}.


\section{Epistemic Uncertainty Analysis}
\label{sec:uncertainty}

The scaling study~(Section~\ref{sec:results_scaling}) characterises translation quality
for each configuration but treats each as a deterministic predictor.
In an unpaired setting no paired SR target supervises the generator, so a model may
produce a stain that appears plausible yet does not reproduce the underlying tissue
structure, and without ground truth such errors cannot be detected per tile.
We can, however, train independently seeded models and ask whether they converge on the
same translation.
To this end, we train deep ensembles of $K{=}10$ independently seeded models per family
and treat the pixel-wise variance across ensemble members as an epistemic uncertainty
signal: low variance indicates the unsupervised objective steers all members toward a
common SR output, while persistent inter-seed disagreement marks inputs where the
objective alone does not pin down a unique answer.
Variance measures epistemic uncertainty, not correctness, and must be read alongside
CPA~MAE from the scaling study: a model can be confidently wrong.

\noindent\textbf{Model selection.}
For each family we select the generator size trained on the full training set (100\%
data fraction) with the lowest CPA~MAE (FID as tiebreak), marked with a star
($\bigstar$) in Figure~\ref{fig:results_overview} and listed in
Table~\ref{tab:best_per_family}.

\noindent\textbf{Ensemble training.}
We use deep ensembles rather than MC~dropout: a calibration study of unpaired image
translation found MC~dropout poorly calibrated in this regime, while deep ensembles
yield well-calibrated estimates with minimal
tuning~\cite{trustworthy_i2i2025, lakshminarayanan2017simple}.
Each selected configuration is re-trained ten times with distinct random seeds;
hyperparameters and training data match the corresponding scaling-study run.
The ensemble size of $K{=}10$ matches convergence analyses showing that calibration
errors and mean epistemic uncertainty both plateau near ten members for biomedical image
segmentation~\cite{siddiqui2024uncertainty} and prior ensemble-based virtual staining
work~\cite{insilicolabeling2023}; additional members yield diminishing returns
at substantial compute cost, as each member requires a full retraining of the
underlying model.

\subsection{Uncertainty Quantification}
\label{sec:uncertainty_quantification}

For each selected configuration, given an H\&E input $x$ and the $K{=}10$
independently trained generators $\{G_k\}_{k=1}^{K}$ (schematic in \suppsec{5}), the ensemble mean is the pixel-wise average
$\bar{y}(x) = \frac{1}{K}\sum_{k} G_k(x)$, and the epistemic uncertainty map
$\mathbf{U}(x) \in \mathbb{R}^{H \times W}$ is the sum of per-channel sample variances
across ensemble members~\cite{lakshminarayanan2017simple}:
\begin{equation}
  \mathbf{U}(x) = \sum_{c=1}^{C} \frac{1}{K-1}
  \sum_{k=1}^{K} \bigl(G_k(x)_{c} - \bar{y}(x)_{c}\bigr)^{2},
  \label{eq:uncertainty}
\end{equation}
where $C{=}3$ is the number of output channels (RGB).
Summing channels preserves the total disagreement signal; high values of
$\mathbf{U}(x)$ flag regions where the model fails to converge on a consistent
translation across seeds, i.e., the inter-seed disagreement signal motivated above.
A per-tile summary is the spatial mean standard deviation
$\bar{\sigma}(x) = \frac{1}{|\mathcal{T}|}\sum_{p \in \mathcal{T}} \sqrt{\mathbf{U}(x)_p}$
over tissue pixels $\mathcal{T}$, used as the per-family epistemic uncertainty
statistic summarised in Figure~\ref{fig:uncertainty_boxplot}.

\subsection{Results: Ensemble Variance}
\label{sec:unc_variance_results}

\begin{figure*}[t!]
  \centering
  \includegraphics[width=0.85\linewidth]{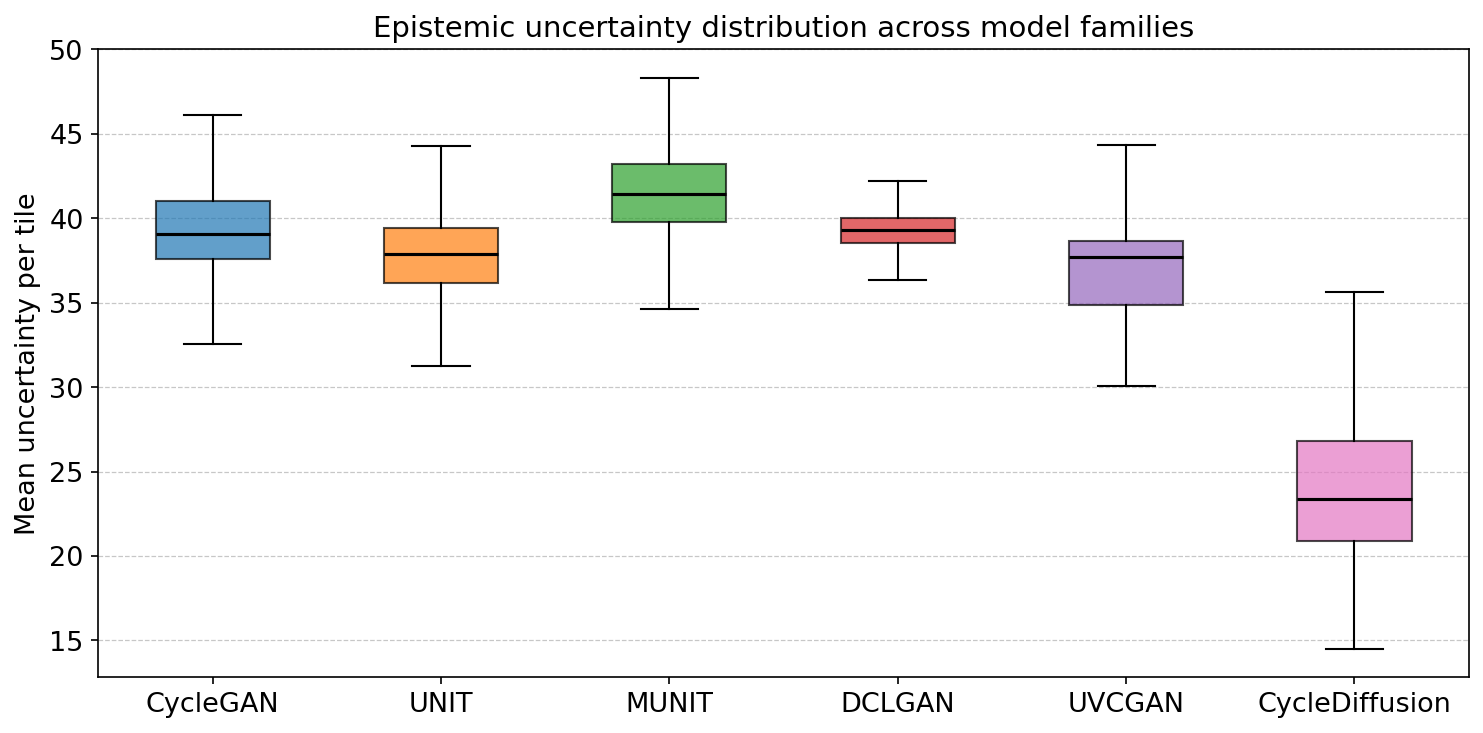}
  \caption{%
    Distribution of per-tile epistemic uncertainty $\bar{\sigma}$ across $9{,}365$
    test tiles (those with at least $0.1\%$ tissue mask coverage) for each of the six
    model families; median values are annotated.
    GAN-based families cluster within a narrow band (median $37.68$--$41.42$,
    IQR $1.48$--$3.80$); CycleDiffusion sits below the cluster (median $23.35$) yet
    shows substantially wider spread (IQR $5.89$).%
  }
  \label{fig:uncertainty_boxplot}
\end{figure*}

Figure~\ref{fig:uncertainty_boxplot} reports the per-tile $\bar{\sigma}$ distribution
across the $9{,}365$ test tiles for each model family.
The reported values are spatial-mean standard deviations summed across the three
RGB channels on the $0$--$255$ intensity scale, so a per-tile value of ${\sim}39$
corresponds to roughly $13$ intensity units per channel (${\sim}5\%$ of the full
range).

\paragraph{GAN families.}
The five GAN-based methods cluster within a narrow uncertainty band, with medians
ranging from $37.68$ (UVCGAN) to $41.42$ (MUNIT) and IQRs of $3.26$--$3.80$.
DCLGAN is the exception within this group: its median ($39.29$) is comparable to the
other GAN families, yet its IQR ($1.48$) is approximately half that of any other GAN
family.
MUNIT records the highest median ($41.42$) among GAN families.

\noindent\textbf{CycleDiffusion.}
CycleDiffusion sits distinctly below the GAN cluster in typical uncertainty (median
$23.35$), yet its IQR ($5.89$) is $1.6$--$4\times$ wider than any GAN family.
On many tiles the ten independently seeded diffusion models converge on a consistent
SR translation, while on a substantial subset they diverge sharply, a tile-level
variability absent in all GAN families.

\noindent\textbf{Joint reading with CPA~MAE.}
Combined with the CPA~MAE values of Section~\ref{sec:results_scaling}, the two axes
separate the families: CycleGAN's medium generator pairs the study-low CPA~MAE
($0.008 \pm 0.007$) with moderate uncertainty (median $39.09$, IQR $3.39$), DCLGAN pairs
low CPA~MAE ($0.028$) with the narrowest spread (IQR $1.48$), and CycleDiffusion pairs
moderate CPA~MAE ($0.051$) with the lowest median but widest spread.
UNIT, MUNIT, and UVCGAN occupy intermediate positions on both axes.

A complementary calibration analysis against cycle-reconstruction error
(\suppsec{5}) shows that GAN-based uncertainty maps moderately predict
tile-level error (across-tile Pearson $\rho = 0.50$--$0.66$), with DCLGAN
($\rho = 0.66$, ECE $0.055$) and MUNIT ($\rho = 0.62$, ECE $0.058$) leading.
CycleDiffusion's uncertainty is uncorrelated with cycle error ($\rho = {-}0.03$),
and no family achieves meaningful within-tile spatial calibration.

%
%

\section{Discussion}
\label{sec:discussion}

The scaling study across 54 configurations and the subsequent ensemble analysis reveal
that the optimal scaling point is model-family-specific and not easily predictable
from perceptual metrics alone, and that ensemble agreement separates the six families
along an axis different from CPA~MAE.

\noindent\textbf{Perceptual saturation and the case for task-specific evaluation.}
Across the $37$ stable GAN configurations, excluding UVCGAN's degraded medium and
large variants, perceptual metrics saturate within a narrow band
(Patch-SSIM $0.298$--$0.349$, LPIPS $0.038$--$0.040$, FID $190$--$253$), with Spearman
rank correlations against CPA~MAE of only $r_s{=}0.63$, $0.64$, and $0.55$.
As a concrete illustration, CycleGAN's medium and large generators at 100\% data are
nearly indistinguishable by Patch-SSIM ($0.334$ vs.\ $0.342$), yet their CPA~MAE
differs by more than an order of magnitude ($0.008$ vs.\ $0.085$).
A perceptual-only selection workflow would treat the two as equivalent and propagate
the 10-fold CPA~MAE gap into any downstream fibrosis staging readout.
Consistent with prior calibration analyses of unpaired
translation~\cite{trustworthy_i2i2025}, these findings confirm that task-specific
evaluation is non-negotiable for virtual staining model selection.
The reader study points the same way: the two configurations that bracket the CPA~MAE
range are separated unanimously by expert readers on both detection and plausibility,
while the perceptual metrics place them far closer together.

\noindent\textbf{Minimum viable configurations across scaling.}
CPA~MAE does not improve smoothly with scale: a few configurations record substantially
lower error than their neighbours, partly because the metric carries high variance over
the $N{=}5$ held-out paired specimens.
When the full training set is available, medium-capacity generators are the most reliable
choice across families.
MUNIT is the most stable family overall (CPA~MAE $0.040$--$0.059$ across all nine
conditions), consistent with AdaIN-based style disentanglement acting as an implicit
regulariser.
DCLGAN's small generator remains stable across all data fractions
(CPA~MAE $0.019$--$0.049$), making it a sound choice when training data is scarce.
CycleGAN and UNIT show non-monotonic data-fraction dependence and are less reliable when
training data is scarce.
The medium-capacity sweet spot may itself reflect overfitting at higher capacity under
this 30-specimen training set, and more diverse data could shift the optimum toward
larger generators; the design cannot test this, since all three data fractions are
nested subsets of one cohort and so cannot separate "more data" from "more diverse
data".

\noindent\textbf{Training instability.}
DCLGAN's large generator collapses at reduced data fractions (FID $327$--$363$,
Patch-SSIM $0.13$--$0.14$) while smaller variants remain stable; we hypothesise this
reflects insufficient patch density for the dual InfoNCE contrastive buffer at lower
data fractions.
UVCGAN's two-stage schedule~\cite{torbunov2022uvcgan} does not scale beyond the small
generator in this data regime; medium and large variants fail across most data fractions
(FID up to $395$, Patch-SSIM as low as $0.14$), possibly because the masked-image
pretraining gains do not transfer to larger generators.
Both failures appear as sharp discontinuities rather than gradual degradation, so
monitoring FID throughout training is essential for early failure detection.

\noindent\textbf{CycleDiffusion as a distinct generative paradigm.}
CycleDiffusion's consistently lower Patch-SSIM ($0.200$--$0.293$ vs.\ $0.298$--$0.342$
for stable GANs) and higher LPIPS reflect the DDIM inversion inductive bias rather
than failure: the noise-space bottleneck enforces domain-invariant structure at the cost
of per-tile pixel sharpness.
The small generator's CPA~MAE rises sharply with data reduction ($0.051$ at 100\%
$\to$ $0.171$ at 25\%), consistent with DDIM inversion requiring well-covered training
distributions to produce stable noise representations.
Counter-intuitively, the medium generator achieves its lowest CPA~MAE at 25\% data
($0.035$ vs.\ $0.082$ at 100\%); one possible explanation is that limited diversity
acts as a regulariser against slide-specific colour artefacts, although we have not
verified this directly.
At 100\% data, the small generator (CPA~MAE $0.051$) is selected for ensemble
uncertainty estimation; the medium generator is preferable under data reduction.

\noindent\textbf{Epistemic uncertainty agreement across families.}
The per-tile mean ensemble standard deviation (Sec.~\ref{sec:unc_variance_results}) separates
the six families along an axis that CPA~MAE alone does not separate.
Median disagreement is modest but perceptible in all families, so the clinically relevant
differences lie in IQR spread and calibration correlation.
Among GAN families, DCLGAN's IQR ($1.48$) is roughly half that of any other GAN
variant, consistent with the dual InfoNCE objective acting as an architectural
regulariser that anchors translations to shared content features across seeds.
Paired with its leading across-tile calibration (Pearson $\rho = 0.66$, ECE $0.055$;
\suppsec{5}), DCLGAN combines low task error with the most informative
ensemble-variance signal among the six families.
CycleGAN's selected medium configuration pairs the study-low CPA~MAE with moderate
ensemble spread, making it the primary candidate when raw task accuracy outweighs
fine-grained uncertainty information.
MUNIT's highest median uncertainty among GAN families ($41.42$) coexists with its most
stable CPA~MAE range, and both follow from style disentanglement: independently seeded
models converge on the same collagen content but sample different style codes via
Adaptive Instance Normalisation, so the SR pixels differ while the fibrosis quantity
does not.
CycleDiffusion's profile is qualitatively different: it records the lowest median
uncertainty ($23.35$) yet the widest IQR ($5.89$).
The shared-noise-space conditioning of DDIM inversion drives ensemble members toward
a near-deterministic output where the conditioning is unambiguous, suppressing
ensemble variation more strongly than any GAN.
On structurally ambiguous tiles the same reverse process diverges sharply, and the
resulting uncertainty signal is uncorrelated with cycle error ($\rho = {-}0.03$).
The diffusion family is therefore best read as conditionally consistent rather than
uniformly more reliable than the GAN alternatives.
These observations hold only under our fixed-budget, from-scratch protocol; we do not
claim diffusion is generally inferior, and pre-trained or noise-inversion variants could
change the picture substantially.

\noindent\textbf{Toward a multi-axis evaluation protocol.}
The three measurement axes used here, perceptual quality, task-specific error, and
ensemble uncertainty, each miss what the others catch: ensemble uncertainty alone says
nothing about whether the consistent answer is correct, and CPA~MAE alone treats
families with very different ensemble agreement as equivalent.
Future virtual-staining benchmarks should therefore report all three jointly, with model
selection driven by the axis closest to the intended downstream use.

\noindent\textbf{Limitations.}
The dataset covers a single tissue type, disease model, and staining laboratory; while
the WSI scanning pipeline is well-standardised (reducing one common source of
cross-centre variability), generalisation to other tissue preparations and disease
contexts remains to be validated.
The bile-duct ligation cohort is a preclinical mouse model, and fibrosis patterns,
collagen architecture, and staining behaviour in human NASH or NAFLD biopsies may
differ in ways that affect ranking transfer.
The held-out test set of $N{=}5$ paired specimens provides limited statistical power:
configurations with CPA~MAE $0.008$ carry a standard deviation of $0.007$, so
configurations that appear substantially lower than their neighbours may reflect
favourable draws rather than reliable improvements, and apparent ranking differences
between closely-spaced configurations should be treated as inconclusive.
Our conclusions therefore rest on aggregate patterns across configurations and on
family-level guidance rather than any single per-configuration ranking.
The reader study is likewise limited to two configurations and three readers, which
brackets the study range but does not support inter-rater statistics.
Patch-SSIM and LPIPS are computed on loosely registered adjacent-section pairs
(Section~\ref{sec:metrics}); because the two sections sample different planes of the
3D tissue, pixel-exact alignment is impossible, so these metrics overstate perceptual
error for translations that are structurally correct but spatially offset.
The uncertainty framework captures epistemic uncertainty only; aleatoric uncertainty
from staining variability and biological heterogeneity between sections is not modelled.

\noindent\textbf{Future work.}
The aleatoric component of uncertainty, dominated by staining variability and
biological heterogeneity between adjacent sections, could be modelled by combining
ensemble disagreement with explicit noise terms that capture these sources directly.
Cross-cohort validation on independent mouse models and on human liver biopsies would
test how the family-level rankings transfer, and may reveal which architectural
inductive biases (cycle reconstruction, contrastive learning, or shared-noise-space
diffusion) generalise beyond the bile-duct ligation preparation.
Pre-trained-prior and noise-inversion approaches (\eg
StainDiffuser~\cite{staindiffuser2024}, InvSR~\cite{yue2024invsr},
F2FLDM~\cite{ho2024f2fldm}) are a natural next axis, testing whether large external
priors improve fidelity beyond the from-scratch, fixed-budget regime studied here.
Evaluation should also broaden to additional downstream, task-specific pathology metrics
beyond CPA~MAE.
Finally, methods that narrow CycleDiffusion's wide ensemble spread (IQR $1.6$--$4\times$
that of any GAN family) could complement its lower median uncertainty for clinical
deployment.

\section{Conclusion}
\label{sec:conclusion}

We benchmarked six unsupervised image-to-image architectures across 54 scaling
configurations on a newly released H\&E~$\to$~Sirius Red mouse liver dataset, jointly
evaluating each configuration on perceptual, distributional, and task-specific axes and
on per-tile epistemic uncertainty from deep ensembles, the first systematic comparison
of epistemic uncertainty across unsupervised stain-to-stain architectures.

For practitioners, the family choice depends on the deployment objective: CycleGAN's
medium generator at full data offers the lowest task-specific error for fibrosis
quantification, DCLGAN's small generator offers the best trade-off between task
accuracy and ensemble agreement when per-tile reliability matters, MUNIT is the safest
default across variable data budgets, and CycleDiffusion is conditionally consistent
rather than uniformly more reliable than the GAN alternatives. More broadly,
task-specific error, perceptual quality, and ensemble uncertainty measure largely
independent axes of model fitness; reporting any single axis gives an incomplete
picture, and future virtual-staining benchmarks should report all three jointly, with
model selection driven by the axis closest to the intended downstream use.

The dataset, tiling pipeline, models, and evaluation code are released as
\textit{I2I-Stain-Zoo} (\url{https://github.com/HoehmeLab/I2I-Stain-Zoo}); the most pressing
next step is cross-cohort validation on independent mouse models and human liver
biopsies.


\section*{Acknowledgements}

This work was supported by the German Federal Ministry of Education and Research (BMBF)
under grants 031L0257J, 031L0256C, 031L0314I, and 031L0313C; by the Deutsche
Forschungsgemeinschaft (DFG) under grant HO4772/5-2; by the European Union under
ARTEMIS (101136299); and by Deutsche Krebshilfe under ARCTIC (70115992).
We further thank Madlen Matz-Soja (Clinic and Polyclinic for Oncology,
Gastroenterology, Hepatology, Pneumology and Infectiology, University Hospital Leipzig)
and Georg Damm (Clinic and Polyclinic for Visceral, Transplantation, Thoracic and
Vascular Surgery, University Hospital Leipzig), the two external expert readers in the
blinded reader study.

\bibliography{references}

\clearpage

\setcounter{section}{0}
\setcounter{figure}{0}
\setcounter{table}{0}
\setcounter{equation}{0}
\renewcommand{\thesection}{S\arabic{section}}
\renewcommand{\thesubsection}{S\arabic{section}.\arabic{subsection}}
\renewcommand{\thefigure}{S\arabic{figure}}
\renewcommand{\thetable}{S\arabic{table}}
\renewcommand{\theequation}{S\arabic{equation}}

\section*{Supplementary Material}


\section{Dataset Details}
\label{sec:supp-dataset}

\paragraph{Source cohort.}
\label{sec:supp-animal_model}

The dataset originates from the bile-duct ligation cohort
of~\cite{ghallab2025asbt}. Briefly, obstructive cholestasis was induced in male
C57BL/6N mice by ligation of the extrahepatic common bile duct~(BDL); sham-operated
controls underwent the identical surgical procedure without ligation. All animal
experiments were approved by the local committee (LANUV, North Rhine-Westphalia,
Germany; application number 81-02.04.2022.A286), as reported
in~\cite{ghallab2025asbt}. The cohort
spans four post-surgical disease stages (day~3, 21, 42, 63) representing early,
intermediate, established, and advanced hepatic fibrosis~\cite{bataller2005liver}.
Both BDL and sham arms received vehicle as part of a larger pharmacological trial;
the pharmacologically treated arms are excluded here to avoid intervention-driven
morphological variation that would confound stain translation, yielding a clean
disease-progression continuum driven solely by BDL severity.
This work uses only the vehicle-treated liver sections stained with H\&E and
Sirius Red, the diagnostic gold standard for hepatic
fibrosis~\cite{huang2013sirius}.
In total, 35 animals contribute the liver sections used in this work, spanning the
four disease stages and both treatment conditions, with four to five animals per
stage and condition.

\paragraph{Histology.}
\label{sec:supp-staining}

Following the protocol of~\cite{ghallab2025asbt}, serial 4\,$\mu$m sections were
cut from paraformaldehyde-fixed paraffin-embedded liver tissue blocks: one section
per block was stained with H\&E and a non-adjacent section with Sirius Red.
Sirius Red binds fibrillar collagens (types I and III)
selectively~\cite{constantine1969}, providing high-specificity collagen
visualisation. Because both sections originate from the same tissue block,
macroscopic morphology is consistent across stains; however, the inter-section
separation precludes pixel-level correspondence.

\paragraph{Digitisation.}
\label{sec:supp-wsi_acquisition}

All slides were digitised at 20-fold magnification, with a native pixel resolution
of $0.221\,\mu$m/pixel, and exported to RGB TIFF for tiling. Slides with severe
scanning artefacts (folds, air bubbles, or scanning failures affecting the majority
of the section) were excluded prior to tiling.

\paragraph{Tiling and Preprocessing.}
\label{sec:supp-dataset_tiling}

Tissue masks were generated for each WSI using a pixel classifier in
QuPath~\cite{bankhead2017qupath}: tissue regions were identified by applying an
intensity threshold on the Eosin channel for H\&E slides and on the second channel
of QuPath's default H-DAB deconvolution vectors for SR slides.
No DAB is present on the Sirius Red sections; the H-DAB vector is used only as a
convenient optical-density channel that separates stained tissue from background.
The masked WSI was tiled into non-overlapping $512{\times}512$\,pixel patches
using a stride equal to the tile size; each patch was subsequently downsampled to
$256{\times}256$\,pixels by bilinear interpolation, yielding an effective pixel
resolution of $0.442\,\mu$m/pixel.
Tiles in which less than 50\% of pixels fell within the tissue mask were discarded
from the training set; no tissue filtering was applied to the test set.
The test specimens are co-registered, so their H\&E and SR sections are tiled on a
single shared grid; with no tissue filtering applied, both domains therefore yield
identical test tile counts by construction.
All retained tiles were normalised channel-wise to $[-1,\,1]$ by subtracting
$0.5$ and dividing by $0.5$ per channel.
This preprocessing was applied identically to both the H\&E~(domain~A) and
SR~(domain~B) tile sets.

Table~\ref{tab:supp-tile_counts} reports the resulting tile counts per domain and split.

\begin{table}[h]
\centering
\caption{Tile counts after tissue filtering. Each paired specimen contributes two
WSIs, one H\&E and one SR. Training subsets (25\%, 50\%) are
strict subsets of the full 100\% training set.}
\label{tab:supp-tile_counts}
\begin{tabular}{lcccc}
\toprule
\textbf{Split} & \textbf{Paired specimens} & \textbf{H\&E tiles} & \textbf{SR tiles} & \textbf{Data fraction} \\
\midrule
Train (100\%) & 30 & 201{,}257 & 196{,}415 & 100\% \\
Train (50\%)  & 15 & 109{,}049 &  96{,}292 & 50\%  \\
Train (25\%)  &  7 &  49{,}312 &  49{,}293 & 25\%  \\
Test          &  5 &  16{,}290 &  16{,}290 & ---   \\
\bottomrule
\end{tabular}
\end{table}

\paragraph{Data Splits and Subsets.}
\label{sec:supp-splits}

The dataset was partitioned at the \textit{animal level} to prevent data leakage:
no WSI from the same animal appears in more than one split.
Splits were stratified by disease stage and treatment group, so that all four
stages and both conditions~(BDL, sham) are represented in every training subset.
The test set is the one exception: the sole day-63 BDL specimen eligible for
holding out could not be co-registered across its H\&E and SR sections and was
therefore excluded, leaving the most advanced fibrosis stage unrepresented among
the held-out BDL specimens (Table~\ref{tab:supp-split_stats}).
The training set comprises 30 paired specimens (60 WSIs) organised into sequentially
numbered per-specimen subdirectories for each domain, enabling reproducible
data-fraction subsets via contiguous folder ranges:
25\%~$(\llbracket 1,7 \rrbracket)$, 50\%~$(\llbracket 1,15 \rrbracket)$, and
100\%~$(\llbracket 1,30 \rrbracket)$.
Smaller subsets are strict subsets of larger ones, ensuring that performance
differences across data fractions are attributable solely to training set size and
not to variation in which slides are included.
The test set comprises 5 held-out paired specimens (10 WSIs) from animals not present
in any training fraction, and is shared identically across all 54 experimental configurations.

Table~\ref{tab:supp-split_stats} summarises the per-split slide counts and disease-stage
distribution.

\begin{table}[h]
\centering
\caption{Dataset split summary. BDL stage indicates the post-ligation day defining
the disease stage; sham = control mice without bile-duct ligation. Each animal
contributes one paired specimen, and each paired specimen contributes two WSIs,
one H\&E and one SR, so the stage columns sum to the specimen count rather than
the WSI count. The test row carries no day-63 specimen: the only candidate could not
be co-registered across its H\&E and SR sections and was excluded.}
\label{tab:supp-split_stats}
\begin{tabular}{lccccccc}
\toprule
\textbf{Split} & \textbf{Specimens} & \textbf{WSIs} &
\multicolumn{5}{c}{\textbf{BDL stage (specimens)}} \\
\cmidrule(lr){4-8}
& & & Sham & Day 3 & Day 21 & Day 42 & Day 63 \\
\midrule
Train & 30 & 60 & 13 & 4 & 4 & 4 & 5 \\
Test  &  5 & 10 & 2 & 1 & 1 & 1 & 0 \\
\bottomrule
\end{tabular}
\end{table}

\paragraph{Public Release.}
\label{sec:supp-public_release}

The original H\&E and Sirius Red WSIs were collected and digitised as part of
the study reported in~\cite{ghallab2025asbt}.
The full release, comprising tissue masks, tile-level metadata, data-split
assignments, tiled image patches, pretrained model checkpoints, and all training
scripts, is distributed under the Creative Commons Attribution~4.0 International
licence (CC~BY~4.0), subject to the data-sharing terms of the cited source study.
The release is hosted as \textit{I2I-Stain-Zoo},
\url{https://github.com/HoehmeLab/I2I-Stain-Zoo}.


\section{Model Details}
\label{sec:supp-models}

\paragraph{CycleGAN.}
\label{sec:supp-cyclegan}

CycleGAN~\cite{zhu2017unpaired} learns two independent generators $G_{A \to B}$ and
$G_{B \to A}$, each a ResNet generator (a $7{\times}7$ convolutional stem followed by
$n_{down}{=}2$ stride-2 downsamples, $n_{\text{blocks}}$ Instance-Normalised residual
blocks at the bottleneck, a symmetric transposed-convolutional decoder, and a Tanh
output)~\cite{zhu2017unpaired}.
Two $70{\times}70$ PatchGAN discriminators~\cite{isola2017image} $D_A$ and $D_B$
distinguish real from generated images in each domain, trained with a Least-Squares
GAN~(LSGAN) objective~\cite{mao2017lsgan} throughout.
Beyond adversarial training, a cycle-consistency loss enforces that translating an image
to the other domain and back recovers the original:
\begin{equation}
  \mathcal{L}_{\text{cyc}} =
    \bigl\| G_{B \to A}(G_{A \to B}(x_A)) - x_A \bigr\|_1 +
    \bigl\| G_{A \to B}(G_{B \to A}(x_B)) - x_B \bigr\|_1,
\end{equation}
weighted by $\lambda_{\text{cyc}}{=}10$.
An optional identity loss ($\lambda_{\text{id}}{=}0.5$) penalises unnecessary colour
shifts when a real image from domain $Y$ is passed through $G_{X \to Y}$.
The total generator objective is:
\begin{equation}
  \mathcal{L}_G = \mathcal{L}_{\text{GAN}} +
    \lambda_{\text{cyc}}\,\mathcal{L}_{\text{cyc}} +
    \lambda_{\text{id}}\,\mathcal{L}_{\text{id}}.
\end{equation}

\paragraph{UNIT.}
\label{sec:supp-unit}

UNIT~\cite{liu2017unsupervised} assumes a shared latent space across domains: images
from both domains are encoded via domain-specific encoders $E_A$ and $E_B$ into a
shared representation from which domain-specific decoders $\text{Dec}_A$ and
$\text{Dec}_B$ reconstruct or translate images.
The shared bottleneck is partitioned into three segments of residual blocks:
domain-private pre-shared blocks ($n_{\text{priv}}^{\text{pre}}{=}3$), shared blocks
($n_{\text{shared}}{=}2$--$3$ depending on size), and domain-private post-shared blocks
($n_{\text{priv}}^{\text{post}}{=}3$), totalling $n_{\text{blocks}}{=}8$--$9$
depending on size.
Stochasticity is introduced via VAE reparameterisation:
$1{\times}1$ convolutions produce per-pixel $\mu$ and $\log\sigma^2$ maps, and a latent
code $z = \mu + \varepsilon\sigma$, $\varepsilon \sim \mathcal{N}(0,I)$, is passed to the
decoders.
The generator loss combines adversarial, cycle-reconstruction L1, and KL-divergence
terms:
\begin{equation}
  \mathcal{L}_G = \lambda_{\text{GAN}}\,\mathcal{L}_{\text{GAN}} +
    \lambda_{\text{recon}}\,\mathcal{L}_{\text{recon}} +
    \lambda_{\text{KL}}\,\mathcal{L}_{\text{KL}},
\end{equation}
with $\lambda_{\text{GAN}}{=}1$, $\lambda_{\text{recon}}{=}10$, and
$\lambda_{\text{KL}}{=}0.01$.

\paragraph{MUNIT.}
\label{sec:supp-munit}

MUNIT~\cite{huang2018multimodal} disentangles each image into a domain-invariant
\textit{content} code $c$ (encoding spatial structure) and a domain-specific
\textit{style} code $s$ (encoding appearance).
Separate content encoders $E^c_A$, $E^c_B$ (ResNet encoders followed by $n_{\text{content}}$
residual blocks) and lightweight style encoders $E^s_A$, $E^s_B$ (three stride-2
convolutions, global average pooling, and a linear layer to a style vector of
dimension $d_s{=}8$) process each domain independently.
Translation is achieved by pairing the content code of the source image with a style
code from the target domain: the style code is injected into the decoder via Adaptive
Instance Normalisation~(AdaIN), in which the affine parameters
$(\gamma,\,\beta)$ are predicted by a style MLP of two fully connected layers
($d_{\text{MLP}}{=}256$) applied to the style vector.
Four AdaIN residual blocks per decoder perform style-conditioned refinement before
standard upsampling.
The generator objective optimises adversarial, image reconstruction, content
consistency, and style consistency terms:
\begin{equation}
  \mathcal{L}_G = \mathcal{L}_{\text{GAN}} +
    \lambda_{\text{img}}\,\mathcal{L}_{\text{img}} +
    \lambda_{\text{c}}\,\mathcal{L}_{\text{content}} +
    \lambda_{\text{s}}\,\mathcal{L}_{\text{style}},
\end{equation}
with $\lambda_{\text{img}}{=}10$, $\lambda_{\text{c}}{=}1$, $\lambda_{\text{s}}{=}1$.

\paragraph{DCLGAN.}
\label{sec:supp-dclgan}

DCLGAN~\cite{han2021dual} augments the CycleGAN framework with a dual contrastive
objective.
Intermediate feature maps are extracted from the encoder output and at three depths of
the residual bottleneck of each generator, giving four feature maps per translation
direction; 256 spatial locations are randomly sampled from each map and projected to a
$256$-dimensional embedding space via per-layer two-layer MLPs.
An InfoNCE loss~\cite{oord2018representation} treats the corresponding anchor--positive
pair (source patch and its translated counterpart) as the positive sample, with all
other patches in the minibatch as negatives, computed symmetrically for both translation
directions:
\begin{equation}
  \mathcal{L}_{\text{DCL}} =
    \mathcal{L}_{\text{NCE}}(A{\to}B) + \mathcal{L}_{\text{NCE}}(B{\to}A),
    \quad \tau{=}0.07.
\end{equation}
The full generator loss is:
\begin{equation}
  \mathcal{L}_G = \mathcal{L}_{\text{GAN}} +
    \lambda_{\text{cyc}}\,\mathcal{L}_{\text{cyc}} +
    \lambda_{\text{DCL}}\,\mathcal{L}_{\text{DCL}},
\end{equation}
with $\lambda_{\text{cyc}}{=}10$ and $\lambda_{\text{DCL}}{=}1$.
Cycle-consistency and the DCL contrastive objective are active together; the
identity loss is disabled ($\lambda_{\text{id}}{=}0$).

\paragraph{UVCGAN.}
\label{sec:supp-uvcgan}

UVCGAN~\cite{torbunov2022uvcgan} replaces CycleGAN's ResNet generator with a
UNet--Vision Transformer hybrid.
A UNet encoder performs $n_{down}{=}4$ stride-2 downsampling steps, collecting skip
connections at each spatial scale (channel progression:
$n_{gf} \to 2n_{gf} \to 2n_{gf} \to 4n_{gf} \to 8n_{gf}$, capped at $8n_{gf}$).
The resulting $16{\times}16$ feature map is flattened into a sequence of tokens and
processed by a lightweight Vision Transformer~(ViT) bottleneck of $n_{\text{ViT}}$
transformer blocks, each consisting of Layer Normalisation, multi-head self-attention,
and a $4{\times}$ feed-forward expansion with GELU activation, combined via ReZero
residual scaling~\cite{bachlechner2021rezero} (all residual scale parameters initialised
to zero).
The UNet decoder restores spatial resolution through symmetric transposed-convolutional
upsampling with skip-connection concatenation.

Training proceeds in two stages.
In \textbf{Stage~1 (pretraining)}, the generator learns to reconstruct images with
randomly masked patches (patch size $32{\times}32$, count 4 per image) using an L1
reconstruction loss on masked regions only, without discriminator involvement.
In \textbf{Stage~2 (finetuning)}, the pretrained generator is used as initialisation
for standard cycle-consistent GAN training ($\lambda_{\text{cyc}}{=}10$,
$\lambda_{\text{id}}{=}0.5$) with both discriminators active.

\paragraph{CycleDiffusion.}
\label{sec:supp-cyclediffusion}

CycleDiffusion~\cite{wu2022cyclediffusion} performs unpaired image-to-image translation
by exploiting an emergent shared Gaussian latent space between two diffusion models trained
simultaneously on their respective domains, without adversarial training or any
cycle-consistency loss term.
Two domain-specific unconditional DDPMs, $\mathcal{M}_A$ for the source domain and
$\mathcal{M}_B$ for the target domain, are each parameterised by a time-conditioned UNet
whose sinusoidal time embedding is projected through a two-layer MLP ($4{\times}$ hidden
expansion) and injected into each residual block via feature-wise affine transformations.
The channel progression is controlled by a base channel count $c_{\text{base}}$ and a
channel-multiplier sequence; self-attention blocks are inserted at the
$16{\times}16$ resolution level.

\textbf{Training.}
Both DDPMs are trained simultaneously in a single training run, each with its own
domain loss and no gradient coupling between them, using the standard epsilon-prediction
objective~\cite{ho2020ddpm}:
\begin{equation}
  \mathcal{L} = \mathcal{L}_{\text{eps}}(A) + \mathcal{L}_{\text{eps}}(B), \qquad
  \mathcal{L}_{\text{eps}} = \mathbb{E}_{x,\,\varepsilon,\,t}
    \bigl[\|\hat{\varepsilon}_\theta(x_t,\,t) - \varepsilon\|_2^2\bigr],
\end{equation}
where $x_t = \sqrt{\bar\alpha_t}\,x_0 + \sqrt{1-\bar\alpha_t}\,\varepsilon$ is the
noisy image at timestep $t$ under a linear noise schedule
($\beta_1{=}10^{-4}$, $\beta_T{=}2{\times}10^{-2}$, $T{=}1000$).

\textbf{Translation.}
At inference, DDIM inversion~\cite{song2021ddim} maps $x_A$ to a shared latent code $z$
via $\mathcal{M}_A$, which $\mathcal{M}_B$ then decodes into the target domain:
\begin{equation}
  z = \mathrm{DDIM\text{-}Invert}_{\mathcal{M}_A}(x_A), \qquad
  \hat{x}_B = \mathrm{DDIM\text{-}Sample}_{\mathcal{M}_B}(z).
\end{equation}
Cycle consistency is an inference-time property of the shared latent structure: because
DDIM inversion is a deterministic left inverse of DDIM sampling, the round-trip
$x_A \to z \to \hat{x}_B \to z' \to \hat{x}_A$ approximately reconstructs $x_A$
without any training signal enforcing it.
Both encoding and decoding use deterministic DDIM ($\eta{=}0$) over $T_{\text{DDIM}}{=}200$
steps, incurring approximately twice the cost of a single DDIM run.

\paragraph{Model Size Configurations.}
\label{sec:supp-model_sizes}

To assess the effect of model capacity, each architecture is trained at three scales.
The key hyperparameters controlling generator size are adjusted per model family
(Table~\ref{tab:supp-model_sizes}); all other architectural choices remain fixed.
Parameter counts refer to the single-direction generator~(A\,$\to$\,B) only; bidirectional
models have approximately twice this count when both generators are included.
For CycleDiffusion, the count covers both the domain-A and domain-B UNet branches
(both are required at inference). Because the two DDPMs are trained independently with
a simple epsilon-prediction objective (no cycle-coupling), per-step training cost is
lower than for cycle-consistent GAN training; the trade-off is reversed at inference,
where translating a single tile requires 200 DDIM inversion steps followed by 200
DDIM sampling steps, making CycleDiffusion substantially slower at test time than the
single forward pass of GAN-based generators.

\begin{table}[h]
\centering
\caption{Generator size configurations. Parameter counts are for the single
A\,$\to$\,B generator direction. ``Shared'' denotes shared bottleneck blocks in UNIT.}
\label{tab:supp-model_sizes}
\resizebox{\textwidth}{!}{%
\begin{tabular}{llccccc}
\toprule
\textbf{Model} & \textbf{Size} & \textbf{Params} & \textbf{$n_{gf}$} &
\textbf{$n_{\text{blocks}}$} & \textbf{Additional hyperparameters} \\
\midrule
\multirow{3}{*}{CycleGAN}
  & Small  & 10.20\,M & 64  & 8  & --- \\
  & Medium & 50.18\,M & 128 & 10 & --- \\
  & Large  & 102.26\,M & 192 & 9  & --- \\
\midrule
\multirow{3}{*}{UNIT}
  & Small  & 10.33\,M & 64  & 8  & shared blocks = 2 \\
  & Medium & 50.71\,M & 128 & 10 & shared blocks = 2 \\
  & Large  & 103.44\,M & 192 & 9  & shared blocks = 3 \\
\midrule
\multirow{3}{*}{MUNIT}
  & Small  & 10.14\,M & 64  & --- & content blocks = 3 \\
  & Medium & 47.64\,M & 128 & --- & content blocks = 5 \\
  & Large  & 94.87\,M & 192 & --- & content blocks = 4 \\
\midrule
\multirow{3}{*}{DCLGAN}
  & Small  & 10.20\,M & 64  & 8  & --- \\
  & Medium & 50.18\,M & 128 & 10 & --- \\
  & Large  & 102.26\,M & 192 & 9  & --- \\
\midrule
\multirow{3}{*}{UVCGAN}
  & Small  & 10.01\,M & 48  & --- & $d_{\text{ViT}}{=}96$,\;$n_{\text{ViT}}{=}6$ \\
  & Medium & 48.28\,M & 96  & --- & $d_{\text{ViT}}{=}384$,\;$n_{\text{ViT}}{=}6$ \\
  & Large  & 96.72\,M & 128 & --- & $d_{\text{ViT}}{=}384$,\;$n_{\text{ViT}}{=}17$ \\
\midrule
\multirow{3}{*}{CycleDiffusion}
  & Small  & 10.68\,M & --- & 1 & $c_{\text{base}}{=}48$,\;mult\,=\,[1,2,2,4] \\
  & Medium & 50.29\,M & --- & 2 & $c_{\text{base}}{=}84$,\;mult\,=\,[1,2,2,4] \\
  & Large  & 100.02\,M & --- & 2 & $c_{\text{base}}{=}128$,\;mult\,=\,[1,2,4] \\
\bottomrule
\end{tabular}%
}
\end{table}


\section{Implementation Details}
\label{sec:supp-implementation}

\paragraph{Shared training hyperparameters.}
All models use the Adam optimiser~\cite{kingma2015adam} with learning rate
$2{\times}10^{-4}$, $\beta_1{=}0.5$, $\beta_2{=}0.999$, and a batch size of~1.
A linear learning-rate decay schedule is applied from the halfway point of training
to zero at the final step; no weight decay is used.
Training is step-based~(one optimiser update per step) rather than epoch-based,
ensuring a fair comparison across data-fraction configurations that differ in the
number of available tiles.

\paragraph{Per-model training steps and schedules.}
\begin{itemize}
  \item \textbf{CycleGAN, UNIT, MUNIT, DCLGAN}: $750{,}000$ steps with a single
    Adam optimiser update alternating between the generator and discriminator.
  \item \textbf{UVCGAN}: Two-stage training.
    Stage~1 (\textit{masked-image pretraining}): $250{,}000$ steps with the
    UNet--ViT generator trained under a masked autoencoding objective using the
    ReZero initialisation strategy.
    Stage~2 (\textit{cycle-consistent finetuning}): $500{,}000$ steps initialised
    from the Stage~1 generator checkpoint, with the full cycle-consistency and
    adversarial losses applied.
  \item \textbf{CycleDiffusion}: $750{,}000$ steps with both the domain-A and
    domain-B DDPMs trained simultaneously in a single run, using the
    $\epsilon$-prediction training objective with a linear noise schedule
    ($T{=}1000$ diffusion steps).
    No fine-tuning phase or post-training step-count adjustment is applied.
\end{itemize}

\paragraph{Loss weights.}
Table~\ref{tab:supp-loss-weights} summarises the loss coefficient used by each
model family.

\begin{table}[h]
\centering
\caption{Loss weights used for each model family. Loss terms vary across
architectures; each row lists only the weights applicable to that family.}
\label{tab:supp-loss-weights}
\begin{tabular}{ll}
\toprule
\textbf{Model} & \textbf{Loss weights} \\
\midrule
CycleGAN       & $\lambda_{\mathrm{cyc}}{=}10$, $\lambda_{\mathrm{id}}{=}0.5$ \\
UNIT           & $\lambda_{\mathrm{GAN}}{=}1$, $\lambda_{\mathrm{recon}}{=}10$, $\lambda_{\mathrm{KL}}{=}0.01$ \\
MUNIT          & $\lambda_{\mathrm{img}}{=}10$, $\lambda_{c}{=}1$, $\lambda_{s}{=}1$ \\
DCLGAN         & $\lambda_{\mathrm{cyc}}{=}10$, $\lambda_{\mathrm{id}}{=}0$, $\lambda_{\mathrm{DCL}}{=}1$ \\
UVCGAN         & $\lambda_{\mathrm{cyc}}{=}10$, $\lambda_{\mathrm{id}}{=}0.5$ \\
CycleDiffusion & $\varepsilon$-prediction objective only (no auxiliary weights) \\
\bottomrule
\end{tabular}
\end{table}

\paragraph{Inference details.}
At test time, all GAN-based models perform a single forward pass through the
trained $G_{A \to B}$ generator per tile.
CycleDiffusion requires two sequential DDIM passes per tile: $\mathcal{M}_A$
encodes the source tile into the shared noise latent via DDIM inversion over
$T_{\mathrm{encode}}{=}200$~steps, and $\mathcal{M}_B$ decodes
it via DDIM sampling over $T_{\mathrm{sample}}{=}200$~steps.
Generated tiles are denormalised from $[-1,1]$ to $[0,1]$ and saved with their
original tile identifiers to enable downstream metric computation.


\section{Full Scaling Study Results}
\label{sec:supp-full-results}

Tables~\ref{tab:supp-scaling-a} and~\ref{tab:supp-scaling-b} report all four
evaluation metrics for each of the 54 experimental configurations
($6~\text{models} \times 3~\text{sizes} \times 3~\text{data fractions}$).
Patch-SSIM, LPIPS, and CPA~MAE are reported as mean~$\pm$~std over the $N{=}5$
held-out paired specimens, matching the main paper; FID is a set-level statistic
with no per-sample variance.
Configurations where the model collapsed (FID~$>300$ and Patch-SSIM~$<0.20$)
are flagged with an asterisk.

\begin{table}[h]
\centering
\caption{Scaling configurations, part~1 of~2: CycleGAN, UNIT, MUNIT. S/M/L =
Small/Medium/Large generator. Data fraction is the percentage of training
specimens used. FID is rounded to the nearest integer here and reported to one
decimal in the main paper. Configurations meeting FID~$>300$ \emph{and}
Patch-SSIM~$<0.20$ are flagged
as collapsed (asterisk); a high FID alone is not sufficient. $\downarrow$ = lower
is better; $\uparrow$ = higher is better.}
\label{tab:supp-scaling-a}
\resizebox{\textwidth}{!}{%
\begin{tabular}{llccccc}
\toprule
\textbf{Model} & \textbf{Size} & \textbf{Data \%} &
\textbf{FID} $\downarrow$ & \textbf{LPIPS} $\downarrow$ &
\textbf{Patch-SSIM} $\uparrow$ & \textbf{CPA~MAE} $\downarrow$ \\
\midrule
\multirow{9}{*}{CycleGAN}
 & S & 25  & $206$ & $0.039 \pm 0.001$ & $0.331 \pm 0.041$ & $0.045 \pm 0.052$ \\
 & S & 50  & $219$ & $0.039 \pm 0.001$ & $0.329 \pm 0.039$ & $0.063 \pm 0.058$ \\
 & S & 100 & $210$ & $0.039 \pm 0.001$ & $0.339 \pm 0.040$ & $0.049 \pm 0.042$ \\
 & M & 25  & $217$ & $0.039 \pm 0.001$ & $0.331 \pm 0.046$ & $0.053 \pm 0.051$ \\
 & M & 50  & $245$ & $0.038 \pm 0.002$ & $0.323 \pm 0.039$ & $0.057 \pm 0.033$ \\
 & M & 100 & $202$ & $0.038 \pm 0.001$ & $0.334 \pm 0.041$ & $0.008 \pm 0.007$ \\
 & L & 25  & $207$ & $0.039 \pm 0.001$ & $0.323 \pm 0.045$ & $0.052 \pm 0.062$ \\
 & L & 50  & $193$ & $0.040 \pm 0.001$ & $0.312 \pm 0.036$ & $0.055 \pm 0.061$ \\
 & L & 100 & $201$ & $0.038 \pm 0.001$ & $0.342 \pm 0.041$ & $0.085 \pm 0.032$ \\
\midrule
\multirow{9}{*}{UNIT}
 & S & 25  & $201$ & $0.040 \pm 0.001$ & $0.321 \pm 0.044$ & $0.049 \pm 0.061$ \\
 & S & 50  & $208$ & $0.038 \pm 0.002$ & $0.338 \pm 0.041$ & $0.017 \pm 0.020$ \\
 & S & 100 & $206$ & $0.038 \pm 0.001$ & $0.324 \pm 0.037$ & $0.046 \pm 0.039$ \\
 & M & 25  & $203$ & $0.040 \pm 0.001$ & $0.316 \pm 0.045$ & $0.050 \pm 0.051$ \\
 & M & 50  & $196$ & $0.039 \pm 0.001$ & $0.326 \pm 0.042$ & $0.047 \pm 0.050$ \\
 & M & 100 & $195$ & $0.038 \pm 0.002$ & $0.333 \pm 0.041$ & $0.037 \pm 0.033$ \\
 & L & 25  & $199$ & $0.039 \pm 0.001$ & $0.333 \pm 0.041$ & $0.024 \pm 0.041$ \\
 & L & 50  & $210$ & $0.038 \pm 0.002$ & $0.334 \pm 0.043$ & $0.023 \pm 0.030$ \\
 & L & 100 & $211$ & $0.039 \pm 0.001$ & $0.298 \pm 0.034$ & $0.043 \pm 0.047$ \\
\midrule
\multirow{9}{*}{MUNIT}
 & S & 25  & $194$ & $0.039 \pm 0.001$ & $0.328 \pm 0.037$ & $0.044 \pm 0.028$ \\
 & S & 50  & $196$ & $0.039 \pm 0.001$ & $0.331 \pm 0.033$ & $0.041 \pm 0.019$ \\
 & S & 100 & $207$ & $0.040 \pm 0.001$ & $0.308 \pm 0.022$ & $0.051 \pm 0.023$ \\
 & M & 25  & $197$ & $0.039 \pm 0.001$ & $0.316 \pm 0.027$ & $0.059 \pm 0.046$ \\
 & M & 50  & $194$ & $0.039 \pm 0.001$ & $0.321 \pm 0.026$ & $0.053 \pm 0.027$ \\
 & M & 100 & $210$ & $0.040 \pm 0.001$ & $0.310 \pm 0.022$ & $0.043 \pm 0.014$ \\
 & L & 25  & $203$ & $0.039 \pm 0.001$ & $0.325 \pm 0.032$ & $0.040 \pm 0.030$ \\
 & L & 50  & $198$ & $0.039 \pm 0.001$ & $0.312 \pm 0.022$ & $0.055 \pm 0.025$ \\
 & L & 100 & $209$ & $0.040 \pm 0.001$ & $0.305 \pm 0.021$ & $0.055 \pm 0.051$ \\
\bottomrule
\end{tabular}%
}
\end{table}

\begin{table}[h]
\centering
\caption{Scaling configurations, part~2 of~2: DCLGAN, UVCGAN, CycleDiffusion.
Notation and conventions as in Table~\ref{tab:supp-scaling-a}.}
\label{tab:supp-scaling-b}
\resizebox{\textwidth}{!}{%
\begin{tabular}{llccccc}
\toprule
\textbf{Model} & \textbf{Size} & \textbf{Data \%} &
\textbf{FID} $\downarrow$ & \textbf{LPIPS} $\downarrow$ &
\textbf{Patch-SSIM} $\uparrow$ & \textbf{CPA~MAE} $\downarrow$ \\
\midrule
\multirow{9}{*}{DCLGAN}
 & S & 25  & $212$ & $0.039 \pm 0.002$ & $0.322 \pm 0.039$ & $0.049 \pm 0.044$ \\
 & S & 50  & $203$ & $0.038 \pm 0.001$ & $0.341 \pm 0.042$ & $0.049 \pm 0.030$ \\
 & S & 100 & $207$ & $0.039 \pm 0.001$ & $0.339 \pm 0.040$ & $0.028 \pm 0.037$ \\
 & M & 25  & $196$ & $0.039 \pm 0.001$ & $0.316 \pm 0.039$ & $0.047 \pm 0.057$ \\
 & M & 50  & $207$ & $0.038 \pm 0.001$ & $0.327 \pm 0.034$ & $0.019 \pm 0.017$ \\
 & M & 100 & $208$ & $0.039 \pm 0.001$ & $0.334 \pm 0.042$ & $0.042 \pm 0.058$ \\
 & L & 25$^\ast$  & $363$ & $0.046 \pm 0.000$ & $0.143 \pm 0.002$ & $0.091 \pm 0.078$ \\
 & L & 50$^\ast$  & $327$ & $0.049 \pm 0.000$ & $0.126 \pm 0.001$ & $0.082 \pm 0.061$ \\
 & L & 100 & $190$ & $0.038 \pm 0.002$ & $0.349 \pm 0.045$ & $0.039 \pm 0.034$ \\
\midrule
\multirow{9}{*}{UVCGAN}
 & S & 25  & $208$ & $0.039 \pm 0.002$ & $0.329 \pm 0.036$ & $0.039 \pm 0.039$ \\
 & S & 50  & $253$ & $0.038 \pm 0.001$ & $0.337 \pm 0.038$ & $0.077 \pm 0.071$ \\
 & S & 100 & $242$ & $0.039 \pm 0.001$ & $0.311 \pm 0.032$ & $0.035 \pm 0.021$ \\
 & M & 25$^\ast$  & $310$ & $0.043 \pm 0.000$ & $0.152 \pm 0.003$ & $0.094 \pm 0.078$ \\
 & M & 50  & $337$ & $0.041 \pm 0.001$ & $0.222 \pm 0.009$ & $0.084 \pm 0.071$ \\
 & M & 100$^\ast$ & $343$ & $0.043 \pm 0.000$ & $0.142 \pm 0.004$ & $0.118 \pm 0.053$ \\
 & L & 25  & $290$ & $0.042 \pm 0.001$ & $0.218 \pm 0.009$ & $0.095 \pm 0.077$ \\
 & L & 50$^\ast$  & $395$ & $0.040 \pm 0.001$ & $0.147 \pm 0.004$ & $0.086 \pm 0.076$ \\
 & L & 100 & $287$ & $0.038 \pm 0.001$ & $0.219 \pm 0.009$ & $0.063 \pm 0.037$ \\
\midrule
\multirow{9}{*}{CycleDiffusion}
 & S & 25  & $229$ & $0.044 \pm 0.000$ & $0.200 \pm 0.006$ & $0.171 \pm 0.045$ \\
 & S & 50  & $224$ & $0.044 \pm 0.000$ & $0.221 \pm 0.012$ & $0.158 \pm 0.053$ \\
 & S & 100 & $210$ & $0.044 \pm 0.000$ & $0.217 \pm 0.009$ & $0.051 \pm 0.029$ \\
 & M & 25  & $208$ & $0.042 \pm 0.000$ & $0.255 \pm 0.020$ & $0.035 \pm 0.017$ \\
 & M & 50  & $209$ & $0.042 \pm 0.001$ & $0.276 \pm 0.032$ & $0.043 \pm 0.028$ \\
 & M & 100 & $218$ & $0.042 \pm 0.000$ & $0.245 \pm 0.022$ & $0.082 \pm 0.071$ \\
 & L & 25  & $219$ & $0.041 \pm 0.000$ & $0.244 \pm 0.014$ & $0.087 \pm 0.064$ \\
 & L & 50  & $228$ & $0.042 \pm 0.001$ & $0.293 \pm 0.015$ & $0.079 \pm 0.057$ \\
 & L & 100 & $322$ & $0.048 \pm 0.001$ & $0.270 \pm 0.025$ & $0.055 \pm 0.043$ \\
\bottomrule
\end{tabular}%
}
\end{table}

\section{Uncertainty Calibration Analysis}
\label{sec:supp-uncertainty}

\begin{figure*}[t!]
  \centering
  \includegraphics[width=0.65\linewidth]{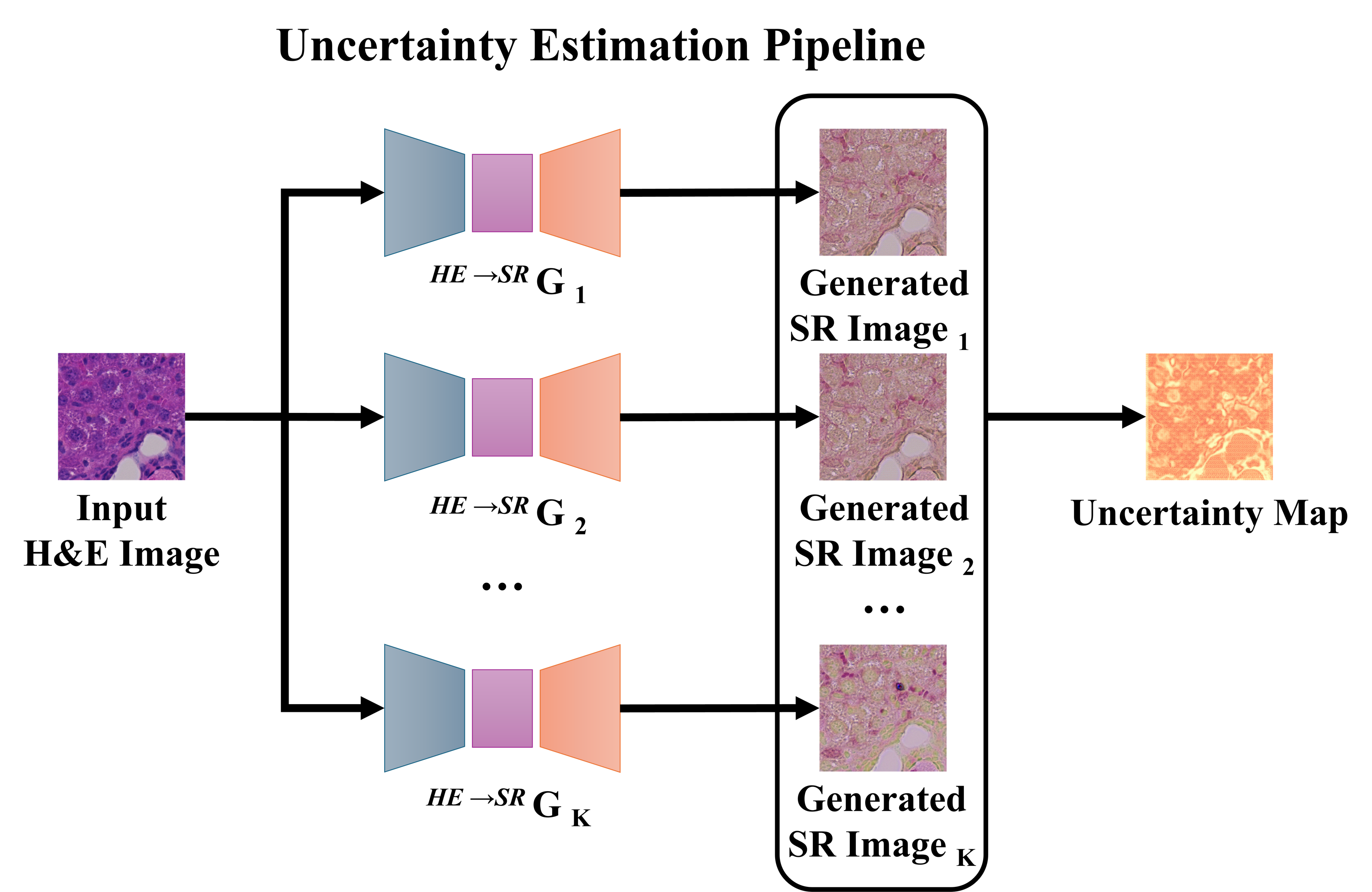}
  \caption{%
    Epistemic uncertainty estimation via deep ensembles.
    An H\&E input tile is forwarded through $K{=}10$ independently trained generators;
    the pixel-wise sample variance across the $K$ SR outputs yields the uncertainty
    map $\mathbf{U}(x)$.%
  }
  \label{fig:supp-uncertainty_maps}
\end{figure*}

\subsection{Uncertainty Quantification}
\label{sec:supp-unc-equations}

Given an ensemble of $K{=}10$ independently trained generators $\{G_k\}_{k=1}^{K}$
for a selected configuration (Figure~\ref{fig:supp-uncertainty_maps}), we derive a
per-tile mean prediction and a spatially resolved epistemic uncertainty map for each
H\&E test tile $x \in \mathcal{D}_{\mathrm{test}}$.

\paragraph{Ensemble mean prediction.}
The ensemble mean image $\bar{y}(x)$ is the pixel-wise arithmetic mean of the ten
generator outputs:
\begin{equation}
  \bar{y}(x) = \frac{1}{K} \sum_{k=1}^{K} G_k(x).
\end{equation}
This mean image is used as the primary virtual stain prediction for all downstream
metric computations.

\paragraph{Epistemic uncertainty map.}
At pixel location $p$, the unbiased sample variance across ensemble members is
computed independently for each colour channel $c \in \{R,G,B\}$:
\begin{equation}
  \sigma^2_c(p) = \frac{1}{K-1} \sum_{k=1}^{K}
    \bigl(G_k(x)_{c, p} - \bar{y}(x)_{c, p}\bigr)^2.
\end{equation}
The three per-channel variance maps are summed to a single-channel scalar
uncertainty map:
\begin{equation}
  \mathbf{U}(x)_p = \sigma^2_R(p) + \sigma^2_G(p) + \sigma^2_B(p),
\end{equation}
giving $\mathbf{U}(x) \in \mathbb{R}^{H \times W}$ where $H{\times}W{=}256{\times}256$
pixels.
The $K{-}1$ denominator gives the unbiased estimator of the population variance under
a finite ensemble~\cite{lakshminarayanan2017simple}.
Summing rather than averaging channels preserves the total disagreement signal: a
pixel where all three channels disagree registers higher uncertainty than one where
only a single channel varies.

\paragraph{Tile-level scalar uncertainty.}
Tile-level uncertainty is the spatial mean standard deviation over tissue pixels
$\mathcal{T}$:
$\bar{\sigma}(x) = \frac{1}{|\mathcal{T}|}\sum_{p \in \mathcal{T}} \sqrt{\mathbf{U}(x)_p}$.
This scalar is used in the across-tile calibration analyses.

\subsection{Calibration Methodology}
\label{sec:supp-unc-calibration}

\paragraph{Error proxy: cycle-reconstruction error.}
In the unpaired setting no pixel-level ground truth exists, as the H\&E and SR
images originate from different physical sections.
We therefore use cycle-reconstruction error as a proxy for per-pixel translation
difficulty.
For a translated tile $B' = G_{A \to B}(A)$, the cycle-reconstruction error at
pixel $(x, y)$ is:
\begin{equation}
  E(x, y) = \bigl|A(x, y) - G_{B \to A}(B')(x, y)\bigr|,
\end{equation}
where $G_{B \to A}\colon B \to A$ is each model's own inverse generator applied to
the forward translation $B'$.
Since all six model families train paired forward and inverse generators, the
self-cycle is available without any external component.
All calibration results in this work are stated as ``calibration with respect to
cycle-reconstruction error''; they do not measure calibration against true
translation error, which is unavailable.

\paragraph{Tile preparation and tissue masking.}
Calibration is computed on a version of the test set re-tiled with zero overlap
(stride equal to tile size, $256{\times}256$ pixels) to avoid double-counting
border pixels under overlapping tiling.
Tiles with at least $0.1\%$ tissue mask coverage are retained, yielding
$N{=}9{,}365$ tiles across the 5 test WSIs; predominantly-background tiles are
excluded.
All per-pixel metrics are restricted to tissue pixels identified by the
preprocessing tissue mask; background pixels, where both $U$ and $E$ are
near-zero, are excluded to prevent spurious inflation of rank correlations.

\paragraph{Spearman $r_s$ (spatial calibration) within-tile.}
For each tile, $U$ and $E$ are flattened over tissue pixels and their Spearman rank
correlation $r_s$ is computed.
A value near $+1$ indicates that high-uncertainty pixels co-locate with high-error
pixels within the tile; near $0$ means the spatial structure of uncertainty is
uninformative; a negative value indicates anti-calibration.
We report mean~$\pm$~std of $r_s$ across all test tiles as the headline spatial
calibration figure.
Spearman rank correlation is preferred over Pearson because $U$ and $E$ are on
different scales and only their ordinal relationship is of interest.

\paragraph{Spearman $r_s$ and Pearson $\rho$ across-tile.}
Each tile is reduced to two scalars, $\overline{U}$ and $\overline{E}$, the
spatial means over tissue pixels; both Spearman and Pearson correlations are then
computed across all tiles in the test set.
This catches the failure mode where a model is spatially well-calibrated within each
tile but assigns identical mean uncertainty to all tiles, making it useless for
flagging unreliable tiles for pathologist review.
Both coefficients are reported because tile-mean values tend to be more linearly
related than per-pixel values, making Pearson a complementary diagnostic.

\paragraph{ECE and reliability diagram tile-level.}
$U$ and $E$ are jointly rescaled to $[0,1]$ using dataset-global $p_1$--$p_{99}$
percentile bounds, making the diagonal of the reliability diagram a meaningful
reference and ECE scores directly comparable across architectures within this
dataset.

Each tile is represented by scalars $\overline{U_n}$ and $\overline{E_n}$ (spatial
means of the normalised maps over tissue pixels) and tiles are binned by
$\overline{U_n}$-quantile into $B{=}10$ equal-mass bins.
Within each bin $b$, the mean normalised uncertainty $\overline{U_n}_b$ and mean
normalised error $\overline{E_n}_b$ are computed and plotted against each other;
perfect calibration corresponds to the identity diagonal.
The tile-level ECE is:
\begin{equation}
  \mathrm{ECE}_{\mathrm{tile}} = \sum_{b=1}^{B} \frac{m_b}{M}
    \bigl|\overline{U_n}_b - \overline{E_n}_b\bigr|,
\end{equation}
where $m_b$ is the number of tiles in bin $b$ and $M$ is the total number of test
tiles.
This adapts the classification ECE of~\cite{guo2017ece} to the regression setting;
values are not directly comparable to ECE from classification contexts.

\paragraph{Calibration caveats.}
The three calibration metrics characterise the same underlying property from
complementary angles and should be read as a converging picture rather than
independent evidence.
Cycle-reconstruction error is a proxy, not a ground truth: self-cycle error can be
underestimated when the forward and inverse generators share systematic biases, in
which case both directions may ignore the same tissue feature and the round trip
still reconstructs the source faithfully despite a poor forward translation.
Cross-section biological differences between the H\&E and SR sections add an
irreducible noise floor to ECE that no model can overcome.

\subsection{Per-Model Uncertainty and Calibration Results}
\label{sec:supp-unc-results}

\begin{figure}[t!]
  \centering
  \includegraphics[width=\linewidth]{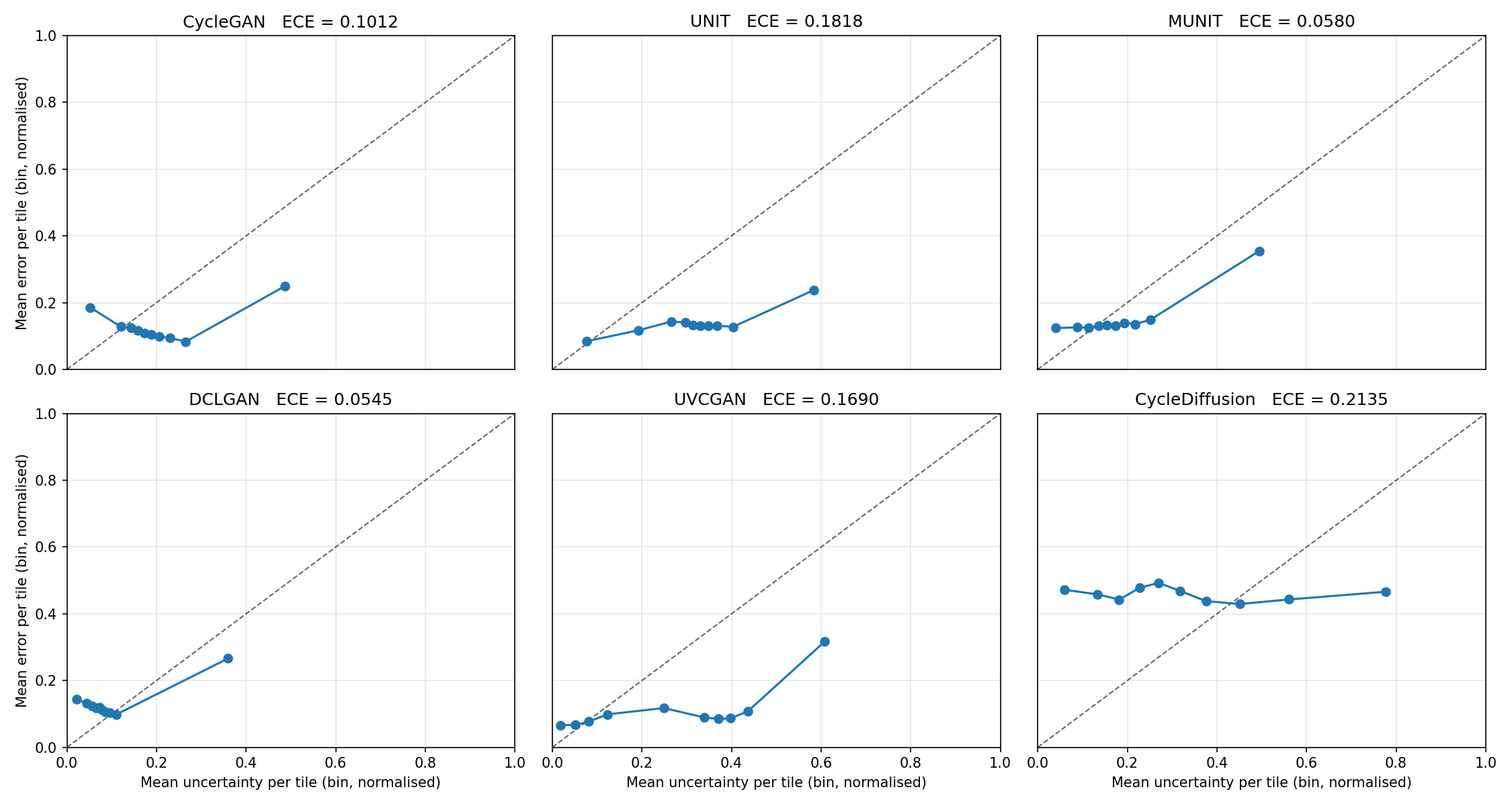}
  \caption{Tile-level reliability diagrams for all six model families (top row:
    CycleGAN, UNIT, MUNIT; bottom row: DCLGAN, UVCGAN, CycleDiffusion).
    Each panel plots mean normalised cycle-reconstruction error per uncertainty bin
    ($y$-axis) against mean normalised uncertainty ($x$-axis); the dashed diagonal is
    the identity line.
    ECE is annotated in each panel title.
    Reliability curves lying below the identity indicate that ensembles assign higher
    uncertainty than the cycle-reconstruction error warrants; the lowest-uncertainty bin
    is an exception in every family, where error exceeds uncertainty.
    CycleDiffusion shows a near-flat error line across all bins, indicating
    uninformative uncertainty.}
  \label{fig:calibration_reliability}
\end{figure}

Tile-level calibration of the per-pixel uncertainty maps against cycle-reconstruction
error, evaluated on the 5 held-out H\&E WSIs. Of their $16{,}290$ test tiles,
$N{=}9{,}365$ contain at least $0.1\%$ tissue-mask coverage and enter the
calibration analysis; the remainder are background and carry no meaningful
uncertainty or reconstruction signal.
Table~\ref{tab:calibration} summarises within-tile Spearman, across-tile Pearson and
Spearman, and ECE for the six selected configurations;
Figure~\ref{fig:calibration_reliability} shows the corresponding reliability diagrams.

\begin{table}[h!]
\centering
\caption{Uncertainty calibration results. Within-tile $r_s$: mean~$\pm$~std of
  per-tile Spearman rank correlation between pixel-level uncertainty $U$ and
  cycle-reconstruction error $E$ over tissue pixels.
  Across-tile Pearson and Spearman: correlations between tile-mean $\overline{U}$
  and tile-mean $\overline{E}$.
  ECE: tile-level expected calibration error after joint $p_1$--$p_{99}$ normalisation.
  $\uparrow$ = higher is better; $\downarrow$ = lower is better.
  All results on $N{=}9{,}365$ tiles from the 5 held-out H\&E WSIs.}
\label{tab:calibration}
\resizebox{\linewidth}{!}{%
\begin{tabular}{llcccc}
\toprule
\textbf{Model} & \textbf{Size} &
\textbf{Within-tile $r_s$} $\uparrow$ &
\textbf{Across-tile Pearson} $\uparrow$ &
\textbf{Across-tile Spearman} $\uparrow$ &
\textbf{ECE\textsubscript{tile}} $\downarrow$ \\
\midrule
CycleGAN       & M & $0.147 \pm 0.109$ & $0.507$ & $-0.178$ & $0.101$ \\
UNIT           & M & $0.139 \pm 0.088$ & $0.497$ & $ 0.169$ & $0.182$ \\
MUNIT          & M & $0.165 \pm 0.173$ & $0.618$ & $ 0.186$ & $0.058$ \\
DCLGAN         & S & $0.157 \pm 0.084$ & $0.659$ & $-0.016$ & $0.055$ \\
UVCGAN         & S & $0.069 \pm 0.130$ & $0.515$ & $ 0.205$ & $0.169$ \\
CycleDiffusion & S & $0.058 \pm 0.095$ & $-0.025$ & $-0.038$ & $0.214$ \\
\bottomrule
\end{tabular}%
}
\end{table}

\begin{figure}[t!]
  \centering
  \includegraphics[width=\linewidth, height=0.82\textheight, keepaspectratio]{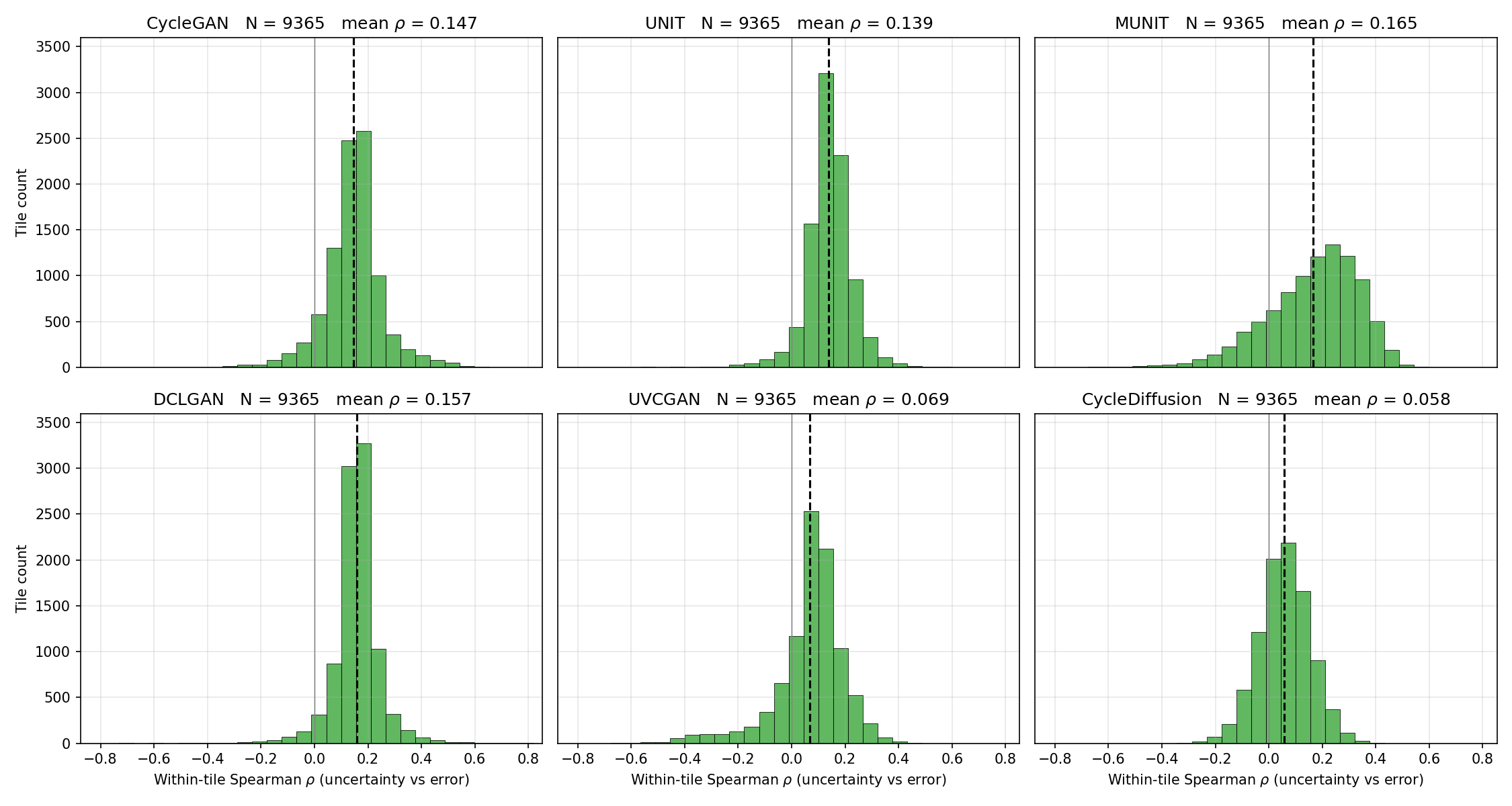}
  \caption{Distribution of within-tile Spearman~$r_s$ between pixel-level uncertainty
    $U$ and cycle-reconstruction error $E$ across the $N{=}9{,}365$ test tiles for each
    model family (top row: CycleGAN, UNIT, MUNIT; bottom row: DCLGAN, UVCGAN,
    CycleDiffusion).
    The dashed vertical line marks the mean~$r_s$.
    GAN-based families peak around $r_s \approx 0.15$--$0.17$; UVCGAN and
    CycleDiffusion are centred near zero, confirming near-absent spatial calibration.
    The broad spread in all panels reflects high tile-to-tile variability.}
  \label{fig:calibration_spearman}
\end{figure}

\textbf{Within-tile spatial calibration.}
All six families produce weak within-tile Spearman correlations ($r_s = 0.058$--$0.165$),
indicating that high-uncertainty pixels are only loosely co-located with high-error
pixels within a tile.
MUNIT ($r_s = 0.165 \pm 0.173$) and DCLGAN ($r_s = 0.157 \pm 0.084$) are the
strongest, though MUNIT's large standard deviation reflects marked tile-to-tile
inconsistency.
UVCGAN ($r_s = 0.069$) and CycleDiffusion ($r_s = 0.058$) show near-zero spatial
calibration.
Figure~\ref{fig:calibration_spearman} shows the full per-tile $r_s$ distributions
behind these means.
These values rule out pixel-level uncertainty as a reliable quality gate.

\begin{figure}[t!]
  \centering
  \includegraphics[width=\linewidth]{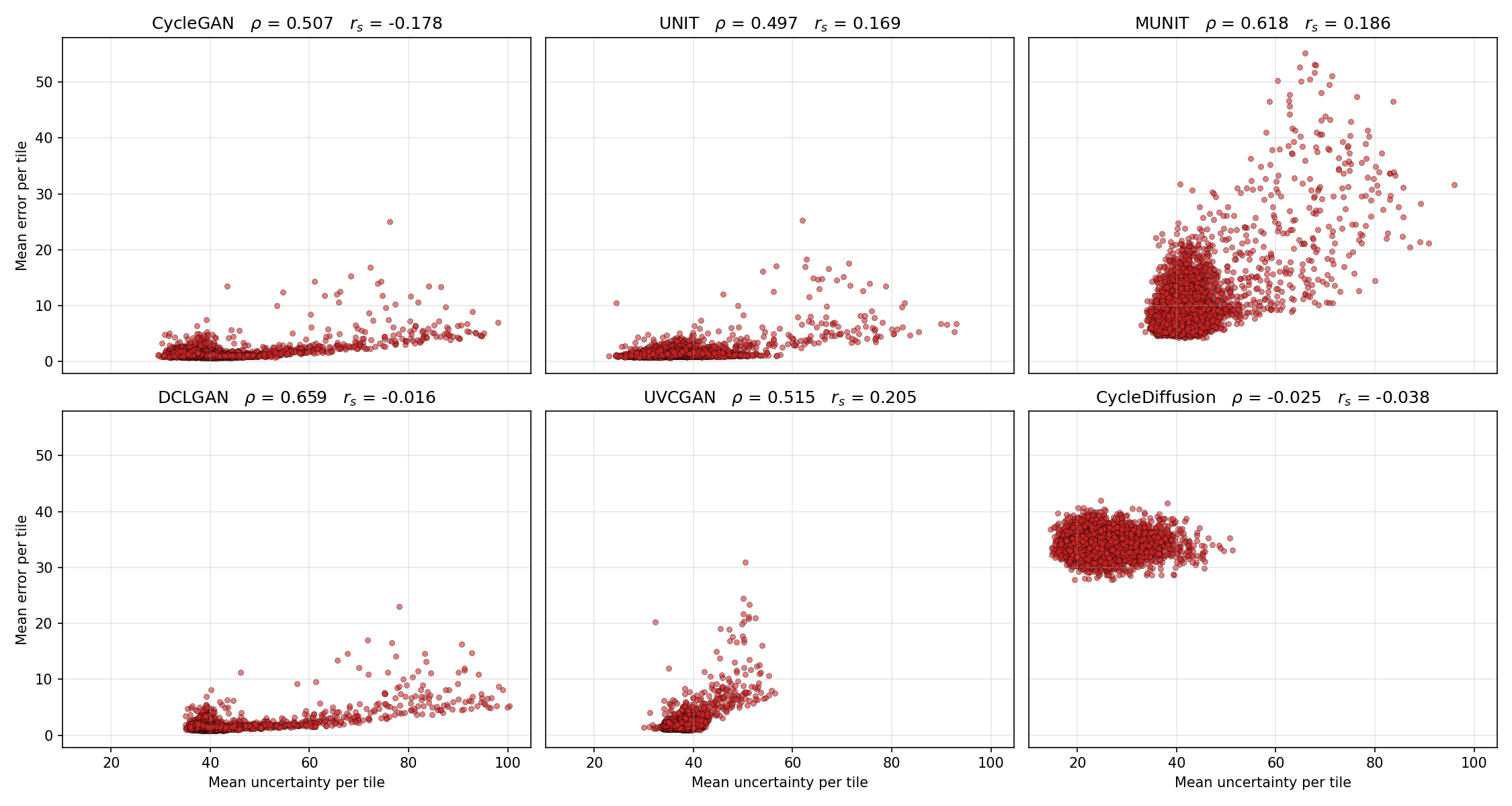}
  \caption{Across-tile scatter plots of tile-mean uncertainty $\overline{U}$ ($x$-axis)
    versus tile-mean cycle-reconstruction error $\overline{E}$ ($y$-axis) for each
    model family (top row: CycleGAN, UNIT, MUNIT; bottom row: DCLGAN, UVCGAN,
    CycleDiffusion).
    Uncertainty is on the same linear $0$--$255$ intensity scale as the per-tile
    $\bar{\sigma}$ of Figure~4 in the main paper.
    Pearson~$\rho$ and Spearman~$r_s$ are annotated per panel.
    GAN-based families show moderate Pearson correlations ($0.50$--$0.66$) driven by
    high-leverage tiles at large uncertainty values; Spearman correlations are
    substantially weaker, indicating the relationship is non-monotonic.
    CycleDiffusion (bottom right) shows no meaningful correlation in either metric.}
  \label{fig:calibration_across_tile}
\end{figure}

\textbf{Across-tile calibration.}
GAN-based families exhibit moderate across-tile Pearson correlations between tile-mean
uncertainty and tile-mean error ($\rho = 0.50$--$0.66$;
Figure~\ref{fig:calibration_across_tile}), with DCLGAN and MUNIT ranking
highest.
The across-tile Spearman correlations are substantially weaker and inconsistent
($r_s = {-}0.18$--$0.21$), revealing that the linear covariation is driven by a
small number of high-leverage tiles rather than a robust monotonic ordering.
CycleDiffusion is wholly uncorrelated (Pearson $\rho = {-}0.03$, Spearman $r_s = {-}0.04$).

\textbf{Reliability diagrams and ECE.}
DCLGAN achieves the lowest tile-level ECE ($0.055$; Figure~\ref{fig:calibration_reliability}),
followed by MUNIT ($0.058$); CycleDiffusion is the worst ($0.214$), with UNIT next
($0.182$).
All six reliability curves lie predominantly \emph{below} the identity line: ensembles
assign higher uncertainty than the normalised cycle error warrants.
This systematic bias is partly structural, as all models minimise cycle-reconstruction
loss, compressing the error proxy relative to inter-member disagreement.
The lowest-uncertainty bin is an exception in every family, where
error exceeds uncertainty, confirming that high-confidence tiles are not guaranteed to
be error-free.

\textbf{CycleDiffusion: categorically uninformative uncertainty.}
CycleDiffusion's reliability diagram presents a flat error line across all ten
uncertainty bins ($\overline{E_n} \approx 0.43$--$0.49$, constant), with near-zero
across-tile correlation in both Pearson and Spearman.
The underlying cause is the DDIM inversion cycle error, which is far
larger than for the GAN families (tile-mean $p_1$--$p_{99}$: 30.0--38.6 versus
0.7--6.7 for CycleGAN, UNIT, DCLGAN, and UVCGAN and 4.9--30.6 for MUNIT, in
unnormalised units) and varies so little across tiles that it
cannot be predicted from ensemble variance.
The uncertainty signal therefore carries no information about translation quality for
CycleDiffusion.

\subsection{Interpretation}

GAN-based families with moderate across-tile Pearson correlations
($\rho = 0.50$--$0.66$) suggest that tiles with elevated ensemble disagreement tend
on average to exhibit greater reconstruction difficulty and may be flagged for
pathologist review; DCLGAN and MUNIT, which rank highest in both across-tile Pearson
and ECE, are the most defensible choices if soft uncertainty-based triage is adopted in
practice.
These calibration rankings align only partially with the CPA~MAE rankings from the
scaling study: CycleGAN achieves the study-best CPA~MAE ($0.008$) despite only
moderate calibration scores, underscoring that perceptual quality, task accuracy,
and uncertainty calibration are three independent axes of evaluation that must each
be considered for reliable, uncertainty-aware evaluation.

\subsection{Summary}

GAN-based ensembles produce a weak but non-trivial tile-level triage signal (across-tile
Pearson $\rho = 0.50$--$0.66$), with DCLGAN and MUNIT as the most reliable candidates.
No family achieves meaningful within-tile spatial calibration, and CycleDiffusion
uncertainty is wholly uninformative.
These results indicate that task-specific CPA~MAE evaluation cannot be replaced by
uncertainty screening alone.

\section{Expert Reader Study}
\label{sec:supp-reader-study}

To complement the automated metrics, three expert readers assessed generated Sirius
Red~(SR) tiles in a blinded study conducted on the held-out set only, using no
additional models, training runs, or data.

\paragraph{Readers.}
Two of the readers are clinical liver experts external to the author list and the third is a
co-author experienced with histological images. None was involved in model development
or evaluation, and none had seen the generated tiles before the study. Readers were not
told how many images came from each source.

\paragraph{Configurations assessed.}
The study evaluates the two configurations that bracket the scaling study: the
study-best CycleGAN medium generator at 100\% training data (CPA~MAE $0.008$) and the
study-worst CycleDiffusion small generator at 25\% training data (CPA~MAE $0.171$),
which record the lowest and highest CPA~MAE among all 54 configurations. Tiles are
$256{\times}256$\,pixel at the same magnification used throughout the paper.

\paragraph{Task 1: real versus generated.}
Readers were shown 45 SR tiles, one at a time in random order: 15 drawn from real SR
sections, 15 from the best configuration, and 15 from the worst configuration.
For each tile the reader recorded a binary judgement, real histochemical stain or
AI-generated, based on staining appearance and tissue morphology. Tiles were judged
independently, and the proportion of real to generated images was withheld so that
readers could not calibrate their responses to an expected base rate.

\paragraph{Task 2: plausibility of generated collagen.}
Task~2 was completed after Task~1 and covers 30 cases. Each case presents one H\&E input
together with two generated SR tiles of the same tissue, labelled Image~A and Image~B
and produced by the two configurations above; the assignment of label to configuration
was fixed across cases and never disclosed. Because the genuine SR section is cut from a
non-adjacent plane and therefore does not correspond pixel-for-pixel to the H\&E, the
H\&E input serves as the reference: readers judged whether the collagen distribution is
biologically plausible given the tissue architecture visible in the H\&E, that is,
whether collagen appears where portal tracts, fibrous septa, and peri-vascular regions
would warrant it, with no fabricated collagen over normal parenchyma and no clearly
missing fibrosis. Agreement was recorded on a forced-choice four-point scale with no
neutral option (1~strongly disagree, 2~weakly disagree, 3~weakly agree, 4~strongly
agree), and Image~A and Image~B were rated independently of each other.

\paragraph{Results.}
Table~\ref{tab:reader_study} summarises both tasks. In Task~1 the readers correctly
identified only $38\%$ of best-configuration tiles as generated, against $62\%$ for the
worst configuration: the best configuration is not reliably distinguishable from real
SR, whereas the worst is. In Task~2 all three readers independently rated the best
configuration as plausible and the worst as implausible, with no overlap between the two
configurations for any reader.

\begin{table}[h!]
\centering
\caption{Blinded reader study on the study-best (CycleGAN-M, 100\% data) and study-worst
  (CycleDiffusion-S, 25\% data) configurations. Task~1 reports the fraction of generated
  tiles correctly identified as generated, pooled over the three readers; values close
  to chance indicate that generated tiles are not distinguishable from real SR. Task~2
  reports each reader's mean plausibility rating on the four-point scale.
  $\downarrow$ = lower is better; $\uparrow$ = higher is better.}
\label{tab:reader_study}
\small
\begin{tabular}{lcc}
\toprule
\textbf{Measure} & \textbf{Best config.} & \textbf{Worst config.} \\
\midrule
Task~1: identified as generated (pooled) $\downarrow$ & $38\%$ & $62\%$ \\
\midrule
Task~2: plausibility, reader~1 $\uparrow$ & $2.7$ & $1.1$ \\
Task~2: plausibility, reader~2 $\uparrow$ & $3.2$ & $1.4$ \\
Task~2: plausibility, reader~3 $\uparrow$ & $3.7$ & $1.7$ \\
\midrule
CPA~MAE (Tables~\ref{tab:supp-scaling-a},~\ref{tab:supp-scaling-b}) $\downarrow$ & $0.008$ & $0.171$ \\
\bottomrule
\end{tabular}
\end{table}

\paragraph{Interpretation and limitations.}
The reader separation tracks the CPA~MAE ranking of the two configurations rather than
their perceptual scores, which place them considerably closer together, and it is
obtained independently of the automated pipeline. The study is deliberately small: it
covers two configurations and three readers, which brackets the range spanned by the
scaling study but does not support inter-rater agreement statistics or a per-family
ranking. Extending the protocol to all six families, and to readers at multiple centres,
is left to future work.

\end{document}